\documentclass[sigconf]{acmart}
\usepackage{enumerate}
\usepackage{multirow}
\usepackage{subfigure}
\usepackage[T1]{fontenc}
\usepackage{aecompl}

\def\BibTeX{{\rm B\kern-.05em{\sc i\kern-.025em b}\kern-.08emT\kern-.1667em\lower.7ex\hbox{E}\kern-.125emX}}

\copyrightyear{2020}
\acmYear{2020}
\setcopyright{acmlicensed}

\begin{document}
\fancyhead{}
\title{Knowledge Adaption for Demand Prediction based on Multi-task Memory Neural Network}


\author{Can Li}
\affiliation{%
  \institution{the University of New South Wales}
  \city{Sydney}
  \country{Australia}
}
\email{lchelen1005@gmail.com}

\author{Lei Bai}
\affiliation{%
  \institution{the University of New South Wales}
  \city{Sydney}
  \country{Australia}
}
\email{lei.bai@student.unsw.edu.au}

\author{Wei Liu}
\affiliation{%
  \institution{the University of New South Wales}
  \city{Sydney}
  \country{Australia}}
  \email{wei.liu@unsw.edu.au}
  
\author{Lina Yao}
\affiliation{%
  \institution{the University of New South Wales}
  \city{Sydney}
  \country{Australia}}
\email{lina.yao@unsw.edu.au}

\author{S Travis Waller}
\affiliation{%
  \institution{the University of New South Wales}
  \city{Sydney}
  \country{Australia}}
  \email{s.waller@unsw.edu.au}
\renewcommand{\shortauthors}{Can Li, et al.}

\begin{abstract}
Accurate demand forecasting of different public transport modes (e.g., buses and light rails) is essential for public service operation. However, the development level of various modes often varies significantly, which makes it hard to predict the demand of the modes with insufficient knowledge and sparse station distribution (i.e., station-sparse mode). Intuitively, different public transit modes may exhibit shared demand patterns temporally and spatially in a city. As such, we propose to enhance the demand prediction of station-sparse modes with the data from station-intensive mode and design a \textbf{M}emory-\textbf{A}ugmen\textbf{t}ed M\textbf{u}lti-task \textbf{Re}current Network (\textbf{MATURE}) to derive the transferable demand patterns from each mode and boost the prediction of station-sparse modes through adapting the relevant patterns from the station-intensive mode. Specifically, \textbf{MATURE} comprises three components: 1) a memory-augmented recurrent network for strengthening the ability to capture the long-short term information and storing temporal knowledge of each transit mode; 2) a knowledge adaption module to adapt the relevant knowledge from a station-intensive source to station-sparse sources; 3) a multi-task learning framework to incorporate all the information and forecast the demand of multiple modes jointly. The experimental results on a real-world dataset covering four public transport modes demonstrate that our model can promote the demand forecasting performance for the station-sparse modes.
\end{abstract}



\maketitle

\section{Introduction} \label{sec:intro}

Each public transport mode such as buses, trains, light rails, and ferries often plays an irreplaceable and integral role in the public transport system and operation of the city. How to design efficient and reliable public transport services is a fundamental and critical problem for cities. Estimating travel demand of various modes of public transit is a critical component to addressing the mentioned problem. Better demand forecasting allows one to better accommodate the public transit demand (e.g., provide sufficient service frequency and design stop/station spacing to reduce waiting, crowding, and improve the attractiveness of public transit services) and improve public transit service efficiency (e.g., efficient and effective routing and scheduling of transit services).

\begin{figure}[htbp]
\centering
\subfigure[]{
\label{fig:amount}
\includegraphics[width=.46\linewidth]{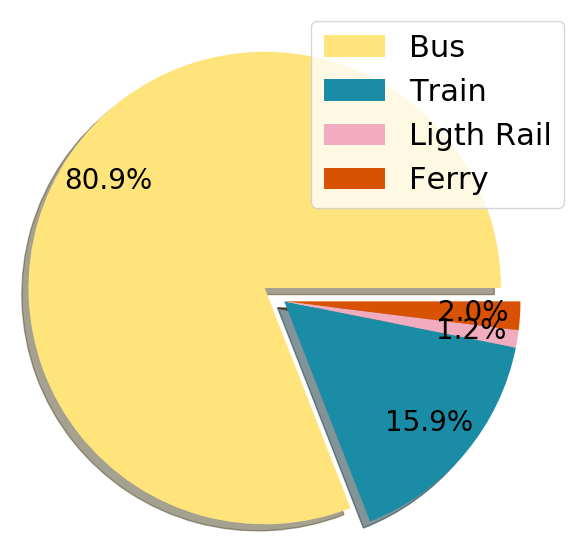}}
\subfigure[]{
\label{fig:geo} 
\includegraphics[width=.46\linewidth]{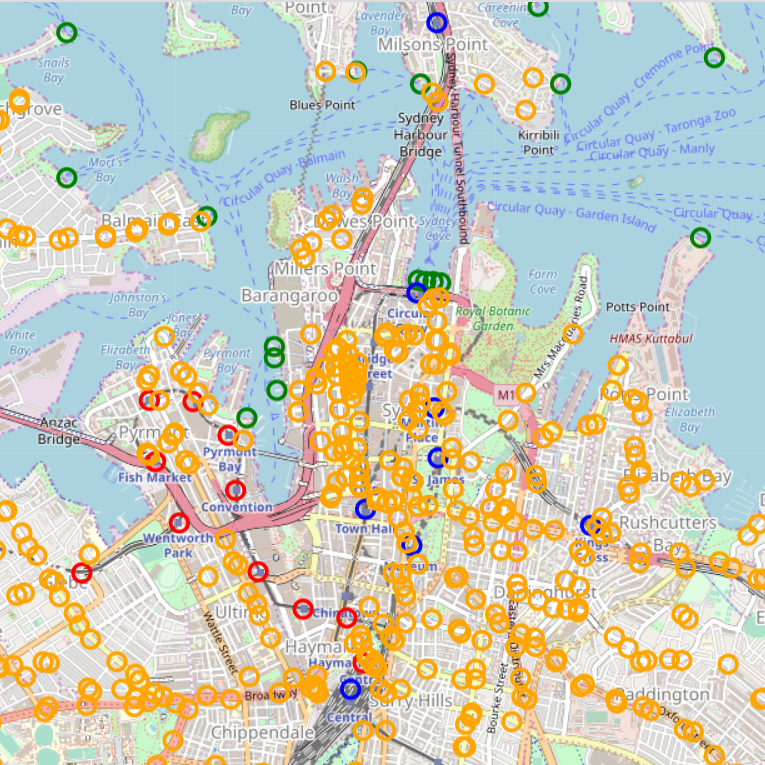}}
\caption{Stations of Four Modes of Public Transport, (a) represents the proportions of different types of public transit stations/stops; (b) represents geographical distribution: color orange represents the bus station; color blue represents the train station; color red represents the light rail station; color green represents the ferry station)}
\label{fig:station} 
\end{figure}

However, the development stages/levels of different public transit modes in a city or region often differ. For instance, Figure~\ref{fig:amount} displays the proportion of stations for the four different modes collected from the Greater Sydney area. The number of bus stations is far more than the other three modes. Moreover, Figure \ref{fig:geo} displays an example of the geographical distribution of some public transportation stations. Consistent with the proportions shown in Figure~\ref{fig:amount}, the covered urban area by the bus stations covers those covered by stations of other public transit modes and the stations of these modes are distributed sparsely. The sparseness of the stations in relation to the train, light rail, and ferry, which often limits the accuracy of demand prediction. From the spatial view, the sparse distribution of stations lacks the features to characterize the local spatial dependencies among the stations. From the semantic view, with limited stations, it is harder to model correlations among stations sharing similar temporal patterns. Thus, we propose to utilize the rich demand data from buses to improve demand prediction for the other three modes considering that demand patterns of different modes may exhibit a certain level of similarity in the same or similar areas. On the one hand, the demand is influenced by the functions of the regions, thus the stations of different transportation modes could share similar demand patterns. On the other hand, near regions may have similar demand patterns \cite{yao2018deep} which leads to analogous demand trends for various modes. This indeed inspires us to utilize the massive data from station-intensive modes (e.g., buses) to improve the demand prediction for station-sparse sources in our work.

In the literature, there has been a long line of studies in traffic data prediction. Traditional methods in the area employ time-series models such as Auto-Regressive Integrated Moving Average (ARIMA) and its variants for traffic flow forecasting \cite{lippi2013short, moreira2013predicting}. These models are less capable to take the non-linear temporal relations into account. In recent years, deep learning models have been explored for capturing non-linear temporal dependencies to forecast the traffic demand. These methods mainly utilized Long-Short Term Memory (LSTM) \cite{bai2019spatio,bai2019passenger,qin2017dual,yao2018deep,wang2018learning} and Gated Recurrent Unit (GRU) \cite{qi2019deep} or temporal convolution module \cite{bai2019stg2seq} to capture the complex non-linear temporal dependencies. However, these methods only focused on one target transport mode with intensive stations/regions. It is hard for these works to handle station-sparse sources for precise prediction. And they were not able to explore the correlations among different transport modes which have great potential to improve the forecasting performance. Although Ye et al. \cite{ye2019co} co-predicted the demand of two transportation modes, their method requires the same level of region coverage for the sources, which lacks the applicability to predict the demand of station-sparse sources. In practice, since the development levels/stages of different cities and modes of transport are uneven, the sparseness of stations/regions for some modes may lead to less accurate and unsatisfactory demand forecasting. Different to all of these existing works, we focus on forecasting the demand of the station-sparse transportation modes and aim to boost the prediction accuracy by considering the demand patterns of different transport modes in one target city and adapting the knowledge learned from the station-intensive source to the station-sparse sources.

To take advantage of the information from a station-intensive source for demand prediction in relation to station-sparse sources, this study proposes \textbf{M}emory-\textbf{A}ugmen\textbf{t}ed M\textbf{u}lti-task \textbf{Re}current Network (\textbf{MATURE}) for station-level demand forecasting in the public transit systems. Specifically, a general multi-task learning framework is constructed for multi-mode demand co-prediction based on LSTM. To capture more accurate temporal correlations and enable the knowledge adaption, we further augment an external memory module to LSTM of each public transit mode for enhancing the ability to capture the long-and-short term information. The extracted temporal knowledge can be stored in the external module and shared by other transport modes that classic LSTM cannot achieve efficiently. After that, utilizing the stored information in the memory module, we design a knowledge adaption mechanism consisting of a boost vector and an eliminate vector to drop the unnecessary features and decide which information should be adapted from the station-intensive transport mode to the station-sparse transport modes. By optimizing the sources jointly, \textbf{MATURE} is capable to adapt appropriate information from the station-intensive mode (e.g., bus) to station-sparse modes (e.g., train, light rail, and ferry) and thus improve the prediction performance of station-sparse transport modes. We validate \textbf{MATURE} on a large-scale real-world dataset collected from the Greater Sydney area including four types of public transport, i.e., bus, train, light rail, and ferry, where the bus contains much intenser information. The comprehensive comparisons with several state-of-the-art methods demonstrate the effectiveness of the proposed model. 

The main contributions of this paper are summarized in the following:

\begin{enumerate}
\item To the best of our knowledge, this is the first study to utilize the data of station-intensive public transit mode to enhance the demand prediction of station-sparse modes in a city, which enlightens a new direction for improving the prediction performance of station-sparse transport modes.
\item This study proposes a novel multi-task deep-learning model - \textbf{MATURE} to improve the prediction performance of the station-sparse transport modes. \textbf{MATURE} can learn more accurate temporal correlations in the historical transit data with the augmented memory modules and adapt the learned knowledge from the station-intensive mode to the station-sparse modes with our knowledge adaption module under the multi-task learning framework.
\item This study conducts extensive experiments on a large-scale real-world public transport dataset collected in a large metropolitan area. The results show that the proposed model significantly outperforms the tested existing methods.
\end{enumerate}

The rest of this paper is organized as follows. First, we introduce the related works in Section~\ref{sec:related work} and define the demand prediction problems in Section~\ref{sec:preliminary}. Then, Section~\ref{sec:mature} presents the proposed \textbf{MATURE} model and technique details. The evaluation of our method and comparison to other methods are presented in Section~\ref{sec:experiments}. Finally, we conclude the paper in Section~\ref{sec:conclusion}.

\section{Related Work} \label{sec:related work}
In this section, we review relevant studies on travel demand forecasting and knowledge adaption methods.

\subsection{Travel Demand Prediction}
The earliest models for demand forecasting are based on traditional time-series regression models such as ARIMA, Kalman Filter, and their variants \cite{lippi2013short,moreira2013predicting,xue2015short}. Lippi et al. \cite{lippi2013short} adopted the combination of the Seasonal ARIMA (SARIMA) model and Kalman filter for good-quality demand predictions. ARIMA and time-varying Poisson models were coupled by Moreira et al. \cite{moreira2013predicting} to predict the spatial distribution of taxi-passenger in a short-term time horizon using streaming data. IMM filter algorithm was applied to individual forecasting models by Xue et al. \cite{xue2015short} for dynamically passenger demand prediction in the next interval. However, these strategies were often hard to capture the non-linear temporal correlations for a precise prediction.

Recently, a number of deep learning methods have shown their success in time-series forecasting such as Fully Connected Layer (FCL), basic RNN, and LSTM \cite{qin2017dual, xu2017real, yi2018deep, huang2019dsanet}. Yi et al. \cite{yi2018deep} proposed a deep neural network-based approach consisting of a spatial transformation component and a deep distributed fusion network for predicting time-series information. Dual-stage attention including an input attention mechanism and a temporal attention mechanism had been added to RNN by Qin et al. \cite{qin2017dual} for capturing the long-term temporal dependencies and forecasting. 

Convolution Neural Network (CNN) \cite{lecun1998gradient} and Graph Convolutional Network (GCN) \cite{kipf2016semi} were also adapted to capture spatial correlations combining with temporal features for better forecasting \cite{seo2018structured, yao2018deep, bai2019stg2seq, zhou2018predicting, wang2019origin}. For example, Li et al. \cite{li2017diffusion} introduced Diffusion Convolutional Recurrent Neural Network (DCRNN) based on a directed graph which captured the spatial dependency by bidirectional random walks for traffic forecasting. An arbitrary graph was structured to the classical RNN by Seo et al. \cite{seo2018structured} to identify spatial structures for sequence prediction. In order to take advantage of the temporal and spatial features, a Deep Multi-View Spatial-Temporal Network (DMVST-Net) was proposed by Yao et al. \cite{yao2018deep} to model correlations among regions sharing similar temporal patterns for taxi demand prediction. Multi-step citywide passenger demands were predicted by Zhou et al. \cite{zhou2018predicting} through an encoder-decoder framework based on convolutional and ConvLSTM units. More recently, Bai et al. \cite{bai2019stg2seq} used a hierarchical graph convolutional structure to capture spatial and temporal correlations simultaneously to predict passenger demand. The grid embedding method for both geographical and semantic neighborhoods was illustrated to capture spatial correlations and then predict origin-destination taxi demand by Wang et al. \cite{wang2019origin}. 

Although the aforementioned works achieved good performance on predicting traffic demand, all of them were based on station-intensive sources with only one forecasting target. They are hard to capture adequate temporal information from station-sparse sources for precise prediction. And they missed the chance to utilize the correlations among different sources for performance improvement. Ye et al. \cite{ye2019co} focused on two transportation modes with a satisfactory level of region coverage which is hard to apply for station-sparse data analysis. Overall, most of the previous works did traffic prediction based on LSTM or RNN which had less capability to store the knowledge from one source which can be adapted to other sources. They also lacked the ability to forecast demand without intensive-station/region data accurately. In practice, due to the unbalanced urban development, the limited data leads to various station-sparse public transportation modes. The works only focused on one domain had less ability to predict the demand of station-sparse sources. Different from the previous works, we aim to adapt the learned knowledge from the station-intensive source mode of public transport to the station-sparse modes for improving their forecasting performance based on the proposed multi-task framework.

\subsection{Knowledge Adaption}
As we discussed in Section~\ref{sec:intro}, it is often hard to provide high-quality demand forecasting of station-sparse sources only based on their own information. Thus, many methods adopted auxiliary data to enhance the performance of prediction \cite{bai2019passenger, xu2017real}. For instance, an end-to-end multi-task model for demand prediction was proposed by Bai et al. \cite{bai2019passenger} where CNN was used to extract spatial correlations and external factors consisting of weather conditions were incorporated to enhance the prediction accuracy. Similarly, the external relevant information including weather and time was applied to LSTM for predicting future taxi requests by Xu et al. \cite{xu2017real}. Unfortunately, the additional data/information is often hard to access (e.g., the public weather data often only contains the overall situation of a city while different regions in the city may have different weather conditions in practice), which certainly limits the applicability of these approaches to a certain extent.

Due to the insufficiency of sources and external data, some works utilized developed previously learning methods to adapt the knowledge to new tasks or domains for prediction \cite{wang2019cross,yao2019learning,wei2016transfer}. An inter-city region matching function was learned by Wang et al. \cite{wang2019cross} to match two similar regions from the source domain to the target domain for crowd flow prediction. For air quality prediction, a Flexible multi-modal transfer Learning (FLORAL) method was proposed by Wei et al. \cite{wei2016transfer} through learning semantically related dictionaries from the source domain and adapting it to the target domain. Yao et al. \cite{yao2019learning} adopted meta-learning to take advantage of the knowledge from multiple cities to increase the stability of transfer for spatial-temporal prediction of the target city. These methods aimed at solving the data-scarce problem lacking training samples (e.g. only has three days data \cite{wang2019cross}) which are different to us. And they did not train the sources jointly which were hard to adapt or share useful knowledge during the training process. Those models designed for data-scarce problems cannot apply to deal with the problem as we described for enhancing the prediction performance for station-sparse sources. 

Compared with the earlier studies, the external factors are not available so they are not incorporated in our work for predicting enhancement. Different from the strategies analyzing the demand patterns of sources with limited training samples, we focus on improving the demand forecasting performance of sources with sparse stations by adapting the useful knowledge learned from the station-intensive source.

\section{Preliminary} \label{sec:preliminary}
In this section, we introduce the dataset used in our work at first. Then we list some mathematical notations and formally define the problem in our work. 

\subsection{Dataset}
The dataset is collected from Sydney covering main public transport services: buses, trains, ferries, and light rails, from 01/Apr/2017 to 30/Jun/2017 including $24$ hours a day covering $6.37$ million users. We choose all lines' information including tap-on and tap-off location (e.g., name, longitude, and latitude of the station), and the number of passengers getting on and off in our experiments. The stations' amount proportion of each public transportation mode is shown in Figure~\ref{fig:amount}. Considering the amount and distribution of each public transportation mode, we use the information of the bus as the station-intensive source and other three public transportation modes (i.e., train, light rail, and ferry) as station-sparse sources. 

\subsection{Problem Formulation}
\noindent \textbf{Demand Series}. For a public transportation mode $D$ (e.g., bus) with $N_D$ stations, we denote the demand of station $i$ at time step $t$ as a scalar $x_t^{D, i}$ which means the amount of passengers during time period $t-1 \thicksim t$. And the unit of the time-step is the duration for counting demand. Next, the demand of all the stations for transport mode $D$ at time-step $t$ can be represented as vector $X^{D}_{t} = \{x^{D,1}_{t}, x^{D,2}_{t}, \cdots, x^{D,i}_{t}, \cdots, x^{D,N_{D}}_{t} \}$. Further, the demand series of transport mode $D$ along time can be denoted as a multivariate time series $\mathbf{X}^{D} = \{X^{D}_{1}, X^{D}_{2}, \cdots,X^{D}_{t}, \cdots, X^{D}_{T} \}$, where $T$ is the total number of time-steps.

\noindent \textbf{Station-level Demand Forecasting Problem}. 
Suppose we have the intensive demand data $\mathbf{X}^{R}$ of mode $R$ and a set of sparse region demand data $\mathbf{X}^{S_{k}}$ of mode $S_{k}$. Given a sequence of demand $\{ X^{R}_{1}, X^{R}_{2}, \cdots, X^{R}_{\tau} \}$ of transport mode $S$, and a sequence of demand $\{ X^{S_{k}}_{1}, X^{S_{k}}_{2}, \cdots, X^{S_{k}}_{\tau} \}$ of station-sparse mode $S_{k}$, our aim is to forecast the demand of each station of station-intensive mode and station-scarce mode in the future time-stamp $T+1$: 
\begin{equation}
    \hat{X}^{R}_{T+1}, \hat{X}^{S_{k}}_{T+1} = \Gamma \{X^{R}_{T-\tau+1}, \cdots, X^{R}_{T-1}, X^{R}_{T},X^{S_{k}}_{T-\tau+1}, \cdots, X^{S_{k}}_{T-1}, X^{S_{k}}_{T} \}
\end{equation}
where $\tau$ is the length of time-steps and $\Gamma(\cdot)$ is the learned prediction function by \textbf{MATURE}. 

\section{Memory-Augmented Multi-task Recurrent Network} \label{sec:mature}

\begin{figure*}[htbp]
    \centering
    \includegraphics[width=.85\linewidth]{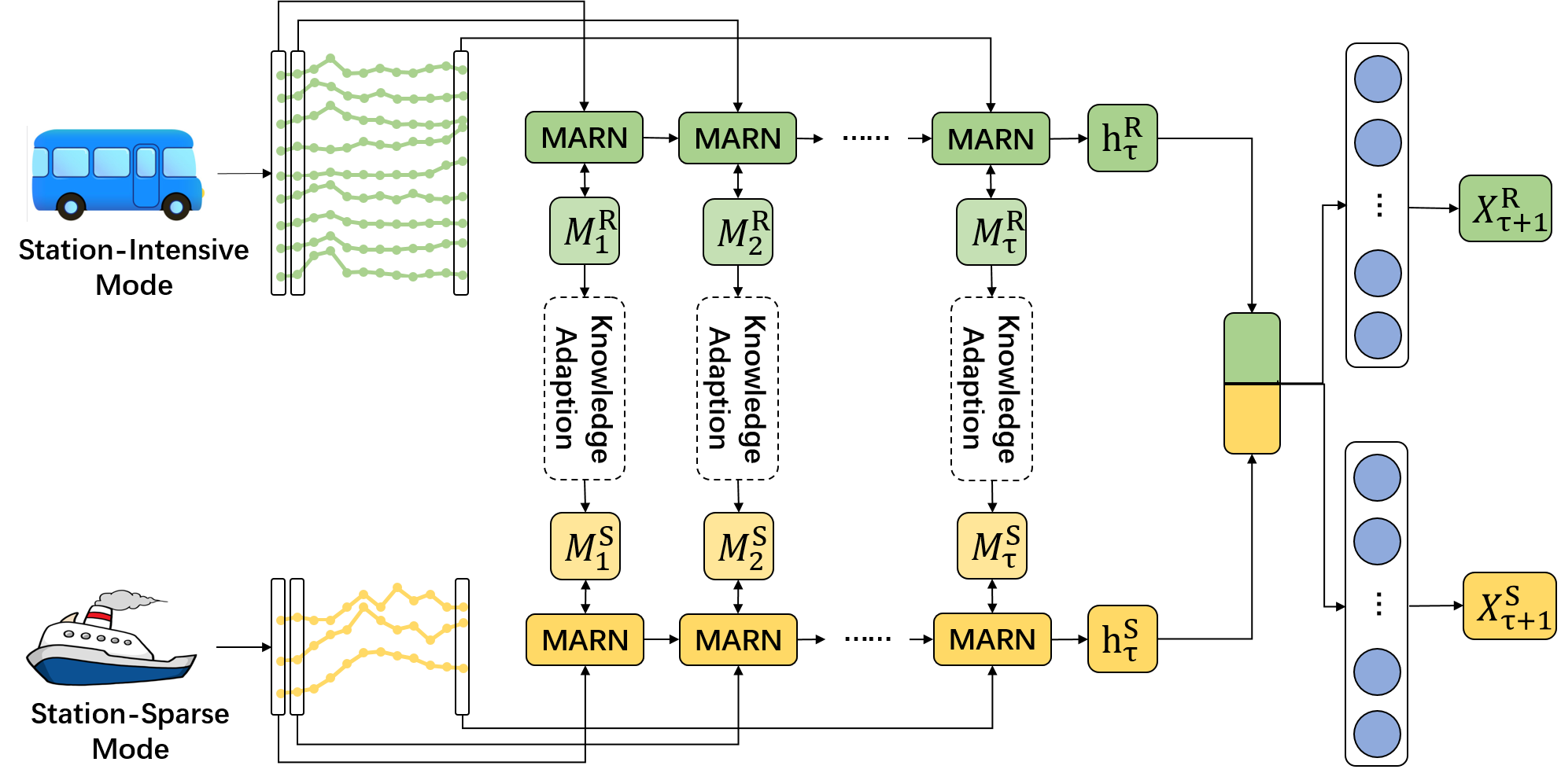}
    \caption{Overall Structure of the Proposed Model: MATURE. $X_{t}$ denotes the demand data, $h_{t}$ represents the hidden state, and $M_{t}$ is the external augmented-memory. $R$ represents the station-intensive source while $S$ represents the station-sparse source. (Use the ferry as an example.)}
    \label{fig:overall} 
\end{figure*}

In this section, we first introduce the basic version of our multi-task framework for enhancing the station-sparse transport mode demand forecasting. Then, we elaborate on how we upgrade the basic framework to the powerful \textbf{MATURE} with two components: Memory-Augmented Recurrent Network (\textbf{MARN}) to enhance the temporal features capturing and store useful information for each transportation mode and Knowledge Adaption Module to incorporate and adapt the knowledge from the station-intensive source to station-sparse sources. The overall structure of the proposed model is shown in Figure~\ref{fig:overall} which will be explained in detail as follows. 

\subsection{Basic Framework} \label{sec:lstm_multi}

To better describe the method for multi-task enhanced forecasting, we describe the basic multi-task learning framework at first.

As described in Section~\ref{sec:intro}, the forecasting performance of the station-sparse sources could be enhanced by the knowledge from the station-intensive source. Thus, a multi-task framework is designed to jointly optimize the station-intensive source and station-sparse source which can learn representations with underlying factors between two sources for further demand forecasting.

To capture the nonlinear temporal relationships, Recurrent Neural Networks have been widely used in NLP tasks to process a sequence of arbitrary length. And nowadays it has been widely used in time-series forecasting \cite{cirstea2018correlated,wang2019memory}. However, for classic RNN, the components of the gradient vector grow or decay exponentially over long sequences \cite{liu2016recurrent}. Therefore, LSTM now becomes a strong tool for time-series forecasting tool due to its capability for learning long-term dependencies and avoiding exploding or vanishing problems by a memory unit and a gate mechanism \cite{qi2019deep,hochreiter1997long}. It is able to learn temporal correlations by maintaining an internal memory cell $\mathbf{c}_{t}$ at time-step $t$. Thus, the data from the station-intensive mode and station-sparse mode are sent into LSTM separately at first to analyze their own temporal correlations. 

In detail, LSTM has an input gate $\mathbf{i}_{t}$, a forget gate $\mathbf{f}_{t}$, an output gate $\mathbf{o}_{t}$, an internal memory cell $\mathbf{c}_{t}$, and a hidden state $\mathbf{h}_{t} \in \mathbb{R}^{m}$. We adopt the LSTM cell in our study as the basic temporal correlations capturing framework which is defined as follows with the input vector $\mathbf{x}_{t} \in \mathbb{R}^{n}$ at the current time-stamp: 

\begin{equation}
\begin{aligned}
    & \mathbf{i}_{t} = \sigma (\mathbf{W}_{i} \mathbf{x}_{t} + \mathbf{U}_{i} \mathbf{h}_{t-1} + \mathbf{b}_{i}) \\
    & \mathbf{f}_{t} = \sigma (\mathbf{W}_{f} \mathbf{x}_{t} + \mathbf{U}_{f} \mathbf{h}_{t-1} + \mathbf{b}_{f}) \\
    & \mathbf{o}_{t} = \sigma (\mathbf{W}_{o} \mathbf{x}_{t} + \mathbf{U}_{o} \mathbf{h}_{t-1} + \mathbf{b}_{o}) \\
    & \mathbf{\theta} _{t} = \tanh (\mathbf{W}_{\theta} \mathbf{x}_{t} + \mathbf{U}_{\theta} \mathbf{h}_{t-1} + \mathbf{b}_{\theta}) \\
    & \mathbf{c}_{t} = \mathbf{f}_{t} \odot \mathbf{c}_{t-1} + \mathbf{i}_{t} \odot \mathbf{\theta} _{t} \\
    & \mathbf{h}_{t} = \mathbf{o}_{t} \odot \tanh (\mathbf{c}_{t})
\end{aligned}
\label{formula:lstm}
\end{equation}
where $\odot$ represents element-wise multiplication, $\sigma$ denotes the logistic sigmoid function $\sigma(u) = 1 \setminus (1 + e^{-u})$, $\mathbf{U}_{i},\mathbf{U}_{f},\mathbf{U}_{o},\mathbf{U}_{\theta} \in \mathbb{R}^{h \times m}$, $\mathbf{W}_{i},\mathbf{W}_{f},\mathbf{W}_{o},\mathbf{W}_{\theta} \in \mathbb{R}^{h \times n}$ are weight matrices, and $\mathbf{b}_{i},\mathbf{b}_{f},\mathbf{b}_{o},\mathbf{b}_{\theta} \in \mathbb{R}^{h}$ are bias vectors.

After extracting temporal features for each source, a simple manner to achieve the knowledge adaption is to merge the data from different sources together. Therefore, the extracted features are concatenated and applied to the output module (e.g., two fully connected layers) to find the relations between the station-intensive source and the station-sparse source for further demand forecasting.

\subsection{Memory-Augmented Recurrent Network} \label{sec:memory}
We now move to introduce the external augmented memory module to extract and store the information from the demand series based on the recurrent neural network. It could enhance the capability to capture the long-and-short term information and share the extracted information with others which the internal memory cell of LSTM cannot achieve efficiently.

\begin{figure}[htbp]
    \centering
    \includegraphics[width=.98\linewidth]{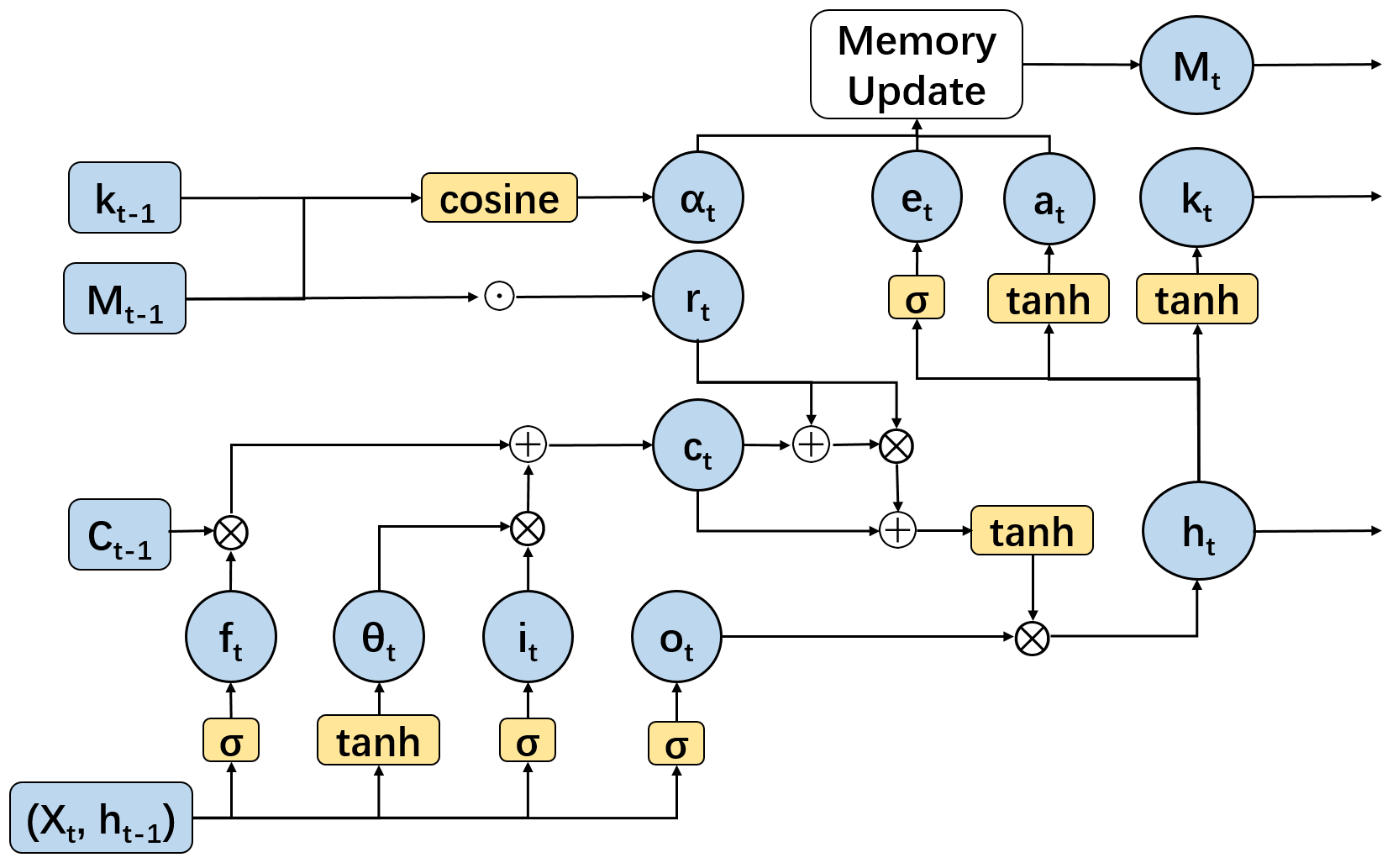}
    \caption{The Architecture of MARN. $\otimes$, $\oplus$, and $\odot$ represents element-wise multiplication, addition, and matrix multiplication respectively. $\sigma$ denotes the logistic sigmoid function. $c_{t}$ represents internal memory cell, $k_{t}$ is a vector emitted by the recurrent network. $b_{t}$ is the boost vector while $l_{t}$ is the eliminate vector, $i_{t}$, $f_{t}$, and $o_{t}$ represent input gate, forget gate, and output gate respectively.}
    \label{fig:structure1} 
\end{figure}

As introduced in Section~\ref{sec:lstm_multi}, the internal memory cell of LSTM has the ability to capture long-and-short term relations for one forecasting task. However, it is hard to be shared by other tasks to improve the forecasting performance for station-sparse sources. In recent years, recurrent neural networks augmented with external memory have been studied with the ability to preserve and share useful information \cite{rae2016scaling}. They are able to learn algorithmic solutions to different complex tasks and were used for language modeling and machine translation \cite{rae2016scaling,liu2016deep}. Motivated by the success of memory mechanism in the NLP area for modeling the long-term temporal correlation, we adopt an external augmented memory module to extract the historical information of each transportation mode and boost the capacity to deal with long-and-short term information. The extracted information can be stored in the augmented memory module and further shared with other transportation modes by the knowledge adaption module, which will be discussed in~\ref{sec:update} for prediction. The architecture of Memory-Augmented Recurrent Network (\textbf{MARN}) is shown in Figure~\ref{fig:structure1}.

In detail, we first define the external augmented memory module as $\mathbf{M}_{t} \in \mathbb{R}^{K \times S}$ at time-step $t$ where $K$ denotes the number of memory segments while $S$ denotes the size of each segment. 

In order to read the useful information from the memory module, we introduce a vector $k_{t}\in \mathbb{R}^{S}$ emitted by LSTM at each time-step $t$ which is defined as:
\begin{equation}
    \mathbf{k}_{t} = \tanh (\mathbf{W}_{k} \mathbf{h}_{t} + \mathbf{b}_{k})
\end{equation}
where $\mathbf{W}_{k},\mathbf{b}_{k}$ represents the weight matrix and bias vector respectively. Then, $\alpha_{t}$ is utilized to decide what information needs to be read from the memory module $\mathbf{M}_{t-1}$ and written to $\mathbf{M}_{t}$ at the next time-step. And $r_{t} \in \mathbb{R}^{S}$ is used for describing the operation of reading effective knowledge from $\mathbf{M}_{t-1}$. The formulas of $\alpha_{t} \in \mathbb{R}^{K}$ and $\mathbf{r}_{t}$ are shown as follows:

\begin{equation}
    \alpha _{t,k} = softmax(f(\mathbf{M}_{t-1,k},\mathbf{k}_{t-1}))
\end{equation}
\begin{equation}
    \mathbf{r}_{t} =  \mathbf{M}_{t-1}^{T} \alpha _{t}
\end{equation}
where $f(\mathbf{x},\mathbf{y})$ is a function to compare the similarity between vector $\mathbf{x}$ and $\mathbf{y}$. And $f(\mathbf{x},\mathbf{y}) = \mathbf{cosine}(x,y)$ where $\mathbf{cosine}$ represents the cosine similarity.

The next step is to update effective information to $\mathbf{M}_{t}$ based on the previous time-step information. Adopting an erase vector $\mathbf{e}_{t} \in \mathbb{R}^{S}$ and an add vector $\mathbf{a}_{t}\in \mathbb{R}^{S}$, we then formulate  $\mathbf{M}_{t}$ as:

\begin{equation}
    \mathbf{M}_{t} = \mathbf{M}_{t-1} \odot (\mathbf{1} - \alpha _{t} \mathbf{e}_{t}^{\mathcal{T}}) + \alpha _{t} \mathbf{a}_{t}^{\mathcal{T}}
\label{formula:memory}
\end{equation}
where
\begin{equation}
\begin{aligned}
    & \mathbf{e}_{t} = \sigma (\mathbf{W}_{e} \mathbf{h}_{t} + \mathbf{b}_{e}) \\
    & \mathbf{a}_{t} = \tanh (\mathbf{W}_{a} \mathbf{h}_{t} + \mathbf{b}_{a})
\end{aligned}
\end{equation}
And $\mathcal{T}$ represents transpose operation, $\mathbf{W}_{e},\mathbf{W}_{a}$ denote the weight matrices while $\mathbf{b}_{e},\mathbf{b}_{a}$ represent bias vectors.

To combine the external augmented memory module $\mathbf{M}_{t}$ computed in Formula~(\ref{formula:memory}) and internal memory cell $\mathbf{c}_{t}$ obtained in Formula~(\ref{formula:lstm}), a deep fusion strategy is introduced instead of simply adding or concatenating to avoid conflicts between the internal memory and external memory. Thus, the hidden state $\mathbf{h}_{t}$ should be formulated as:
\begin{equation}
    \mathbf{h}_{t} = \mathbf{o}_{t} \odot \tanh (\mathbf{c}_{t} + \sigma(\mathbf{W}_{r} \mathbf{r}_{t} + \mathbf{W}_{c} \mathbf{c}_{t}) \odot (\mathbf{W}_{h} \mathbf{r}_{t}))
\end{equation}
where $\mathbf{W}_{r},\mathbf{W}_{c},\mathbf{W}_{h}$ are the weight matrices.

In conclusion, the Memory-Augmented Recurrent Network enhances the ability to handle long-and-short term information which could improve the demand prediction accuracy for all sources. The external memory module also has the strength to store useful knowledge that can be shared by other transportation modes.

\subsection{Knowledge Adaption Module} \label{sec:update}
\begin{figure} [htbp]
    \centering
    \includegraphics[width=.98\linewidth]{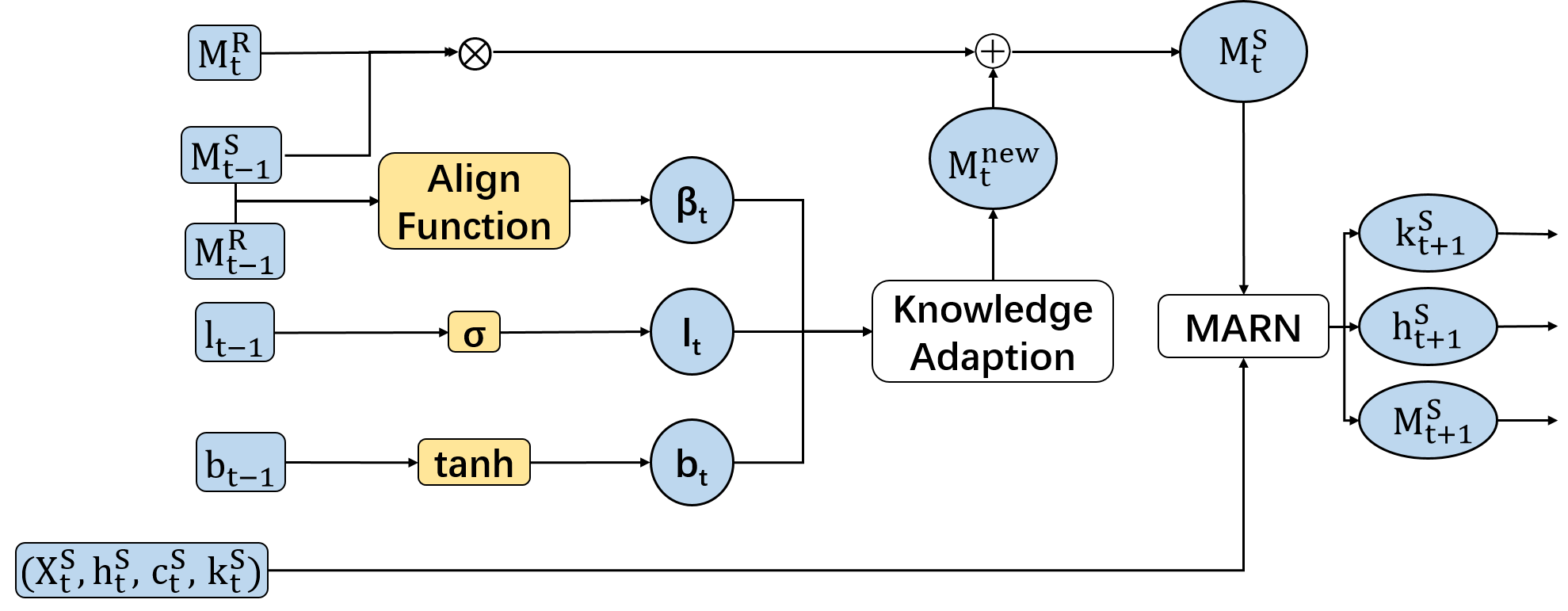}
    \caption{Knowledge Adaption Module}
    \label{fig:structure2} 
\end{figure}

Based on the augmented memory module introduced in Section~\ref{sec:memory}, this section further introduces the proposed knowledge adaption mechanism to adapt useful knowledge from the station-intensive source to station-sparse sources in detail. And the framework of this module is shown in Figure~\ref{fig:structure2}.

Due to the small number of stations as described in Section~\ref{sec:intro}, it is hard to obtain accurate forecasting demand only based on the station-sparse source itself. Thus, we propose to adapt the useful information from the station-intensive source to station-sparse sources for enhancing prediction performance since the extracted knowledge from the station-intensive source could describe the demand patterns in particular areas better which could be useful for station-sparse sources to characterize their own demand patterns in the same area. We adopt the external augmented memory obtained by Formula~(\ref{formula:memory}) of each source. At time-step $t$, we use $\mathbf{M}^{R}_{t}$ and $\mathbf{M}^{S}_{t}$ to denote the external memory of the station-intensive source and the station-sparse source, separately. Directly adapting $\mathbf{M}^{R}_{t}$ to $\mathbf{M}^{S}_{t}$ through concatenation or addition may suffer from feature redundancy for two reasons. On the one hand, it could bring some unnecessary task-specific features to the station-sparse source. On the other hand, some adaptive features may be mixed in the private space. Therefore, we propose a knowledge adaption mechanism to incorporate external memories from different sources.

The first step is to utilize an align function for $\mathbf{M}^{R}_{t}$ and $\mathbf{M}^{S}_{t}$ at time-step $t$ to compare the extracted information from two sources which can be further used to decide what knowledge needed to be adapted. The formula can be shown as:
\begin{equation}
    g(\mathbf{M}^{R}_{t,k},\mathbf{M}^{S}_{t,k}) = \mathbf{v}^{T} \tanh (\mathbf{W}_{g}[\mathbf{M}^{R}_{t,k};\mathbf{M}^{S}_{t,k}])
\end{equation}
where $\mathbf{W}_{g}$ is a parameter matrix and $\mathbf{v}$ denotes a parameter vector. Thus, the corresponding control parameter to decide the amount of knowledge from the station-intensive source should be adapted to the station-sparse source can be computed as:
\begin{equation}
    \beta _{t,k} = softmax(g(\mathbf{M}^{R}_{t-1,k},\mathbf{M}^{S}_{t-1,k})
\end{equation}

After knowing the connection between two sources, we design an adaptive function to fuse the memories $\mathbf{M}^{R}$ and $\mathbf{M}^{S}$ which adopts a boost vector $\mathbf{b}_{t}\in \mathbb{R}^{S}$ and an eliminate vector $\mathbf{l}_{t}\in \mathbb{R}^{S}$. $\mathbf{b}_{t}$ holds the purpose for limiting the task-specific information read from $\mathbf{M}^{R}$ while $\mathbf{l}_{t}$ is used for adapting effective features to the new memory matrix. The function can be written as:
\begin{equation}
    \mathbf{M}^{new}_{t} = \mathbf{M}^{R}_{t-1} (\mathbf{1} - \beta _{t} \mathbf{l}_{t}^{T}) + \beta _{t} \mathbf{b}_{t}^{T}
\end{equation}
where
\begin{equation}
\begin{aligned}
    & \mathbf{b}_{t} = \tanh (\mathbf{W}_{b} \mathbf{b}_{t-1} + \mathbf{b}_{b}) \\
    & \mathbf{l}_{t} = \sigma (\mathbf{W}_{l} \mathbf{l}_{t-1} + \mathbf{b}_{l})
\end{aligned}
\end{equation}

At last, the adaptive matrix of augmented memory for station-sparse source $\mathbf{M}^{S}_{t}$ is computed as:
\begin{equation}
    \mathbf{M}^{S}_{t} = \gamma \times \mathbf{M}^{S}_{t-1} + (1 - \gamma) \times \mathbf{M}^{new}_{t}
\label{formula:12}
\end{equation}
where $\gamma$ is a hyperparameter to decide the proportion of $\mathbf{M}^{new}_{t}$ in the augmented memory. The adaptive information $\mathbf{M}^{S}_{t}$ of station-sparse source then will be sent into the Memory-Augmented Recurrent Network for hidden state capture and further demand prediction.

\subsection{Demand Forecasting and Training Strategy}
Predicting the demand of several public transportation modes independently is hard to adapt useful information from the station-intensive source to the station-sparse source in the training process. Thus, we optimize the demand forecasting of the station-intensive source and the station-sparse source jointly. As shown in Figure~\ref{fig:overall}, the last step of the multi-task learning framework is to concatenate the hidden states of the two sources and then sent them into different fully connected layers for further demand forecasting.

In the training process of \textbf{MATURE}, the objective is to minimize the error between the true demand and the predicted values of the station-intensive source and the station-sparse source simultaneously. The loss function is defined as the mean squared error for time-step length $\tau$, which is formulated as
\begin{equation}
   L(\theta) = \epsilon \times \sum^{T+\tau}_{i=T+1}|| \hat{X}^{R}_{i} - X^{R}_{i} || + (1 - \epsilon) \times \sum^{T+\tau}_{i=T+1}|| \hat{X}^{S}_{i} - X^{S}_{i} ||
\end{equation}
where $\theta$ denotes all the learnable parameters in the proposed \textbf{MATURE} model, $\epsilon$ is a hyperparameter to balance the loss between two sources. Our model can be trained in an end-to-end manner via back-propagation and the Adam optimizer.

\section{Experiments} \label{sec:experiments}
In this section, we first introduce the experiment settings, evaluation metrics, and comparing baselines. In the next, we list the experimental results from two perspectives: overall comparison and ablation study. Then, we discuss the comparison among several public transportation stations with totally different ranges and distribution of demand. The last part mainly focuses on parameter sensitivity (e.g., the hyperparameter $\gamma$ in Formula~(\ref{formula:12})).

\subsection{Experimental Setup}
\textbf{Dataset Setting.} The demand data is normalized by Min-Max normalization for training and re-scaled to the actual values for evaluating the prediction performance. To test the performance of our model, the last $27$ days' data are used for testing while the rest for training and validation. In each experiment, we use one station-sparse source for testing. The unit of time-step we choose is one hour. Since the data volume of the bus is too large which contains some meaningless data (e.g., more than $80\%$ of time-steps in one day have zero demand), we drop the bus stations with an average of fewer than five demands one hour. The number of the bus, train, light rail, and ferry stations are $1573$, $310$, $23$, and $38$ respectively. And we choose the previous $12$ time-steps ($12$ hours) to predict the public transport demand in the next time-step.

\noindent \textbf{Evaluation Metrics.} Two evaluation metrics are used to evaluate the proposed model: Root Mean Square Error (RMSE) and Mean Absolute Error (MAE).

\noindent \textbf{Network Implementation.} The batch size is set to $64$. The proposed model is tuned with the learning rate (from $0.0001$ to $0.01$), the hyperparameter $\epsilon$ in the loss function (from $0$ to $1$ with a step size of $0.1$), $\gamma$ in Formula (\ref{formula:12}) (from $0$ to $1$ with a step size of $0.1$), and the number of hidden units ($128$,$256$,$512$). Different learning rates for different transportation modes are set ($0.002$ for the train, $0.0008$ for the light rail, and $0.0016$ for the ferry). $\epsilon$ is set to $0.1$ for the three modes. The values of $\gamma$ are set to $0.3$ for the train, $0.4$ for the light rail, $0.4$ for the ferry. The weight decay is set to $0.0001$. The number of hidden units is set to $512$. The number of memory segments is $15$ while the size of each segment is $60$. 

\subsection{Baselines} \label{sec:baseline}
We compare the proposed model with the methods in the following:
\begin{itemize}
\item \textbf{Historical Average (HA)}: The predicted demand is computed as the average values of historical demand at the same time interval of every day.
\item \textbf{Linear Regression (LR)}: It models the relations between variables and minimizes the sum of the squares of the errors for prediction.
\item \textbf{eXtreme Gradient Boosting (XGBoost)}: XGBoost was proposed by \citet{chen2016xgboost} based on gradient boosting tree which incorporates the advantages of Bagging integrated learning methods in the evolution process. 
\item \textbf{Multilayer Perceptron (MLP)}: The neural network contains four fulling connected layers and the numbers of hidden units are $256, 128, 128, 64$.
\item \textbf{Long-Short Term Memory (LSTM)}: LSTMs are used to model long- and short-term dependencies and directly applied to predict demand for each transportation mode.
\item \textbf{Graph Convolutional Recurrent Network (GCRN)} \cite{seo2018structured}: It combines CNN on graphs to identify spatial structures and RNN to find temporal patterns for demand forecasting which could improve the forecasting accuracy by simultaneously capturing graph spatial and dynamic information about data.
\item \textbf{Long- and Short-term Time-series network (LSTnet)} \cite{lai2018modeling}: It employs a recurrent-skip network with a convolutional layer to capture the long-term dependence patterns and discover the local dependency patterns for forecasting.
\item \textbf{Dual-stage Attention-based Recurrent Neural Network (DA-RNN)} \cite{qin2017dual}: It has two components: an input attention mechanism to extract relevant driving series at each time-step and a temporal attention mechanism to select relevant encoder hidden states across all time-steps for prediction effectively. The original structure is used for one variant forecasting, we change it to multiple variants forecasting.
\item \textbf{MT-LSTM}: This is the basic multi-task model introduced in Section \ref{sec:lstm_multi}. We use LSTM layers to extract temporal correlations and adopt fully connected layers to analyze the implicit relations for demand forecasting.
\end{itemize}

\begin{table*}[h]
\caption{Overall Comparison between the Proposed Method and Existing Methods}
\begin{tabular}{l|lllll|lllll}
\hline
\multicolumn{1}{c|}{\multirow{2}{*}{\textbf{Method}}} & \multicolumn{5}{c|}{\textbf{MAE}} & \multicolumn{5}{c}{\textbf{RMSE}} \\  \cline{2-11} 
\multicolumn{1}{c|}{} & \textbf{Bus} & \textbf{Train} & \textbf{Light Rail} & \textbf{Ferry} & \textbf{Average} & \textbf{Bus} & \textbf{Train} & \textbf{Light Rail} & \textbf{Ferry} & \textbf{Average} \\ \hline
\textbf{HA} & 13.8485 & 57.3426 & \multicolumn{1}{c}{18.4351} & 28.9801 & 29.6516 & 15.2267 & 63.2433 & \multicolumn{1}{c}{23.4027} & 35.5621 & 34.3587 \\
\textbf{LR} & 15.8681 & 60.0130 & \multicolumn{1}{c}{20.1218} & 26.4974 & 30.6251 & 22.7254 & 103.7374 & \multicolumn{1}{c}{30.9286} & 41.4833 & 49.7187 \\
\textbf{XGBoost}\cite{chen2016xgboost} & 12.6638 & 21.7030 & \multicolumn{1}{c}{12.2439} & 18.3773 & 16.2470 & 17.2568 & 35.0718 & \multicolumn{1}{c}{19.6542} & 28.7830 & 25.1915 \\
\textbf{MLP} & 11.4356 & 25.9388 & \multicolumn{1}{c}{12.0386} & 18.1604 & 16.8934 & 15.5779 & 38.2704 & \multicolumn{1}{c}{18.4082} & 28.8720 & 25.2821 \\
\textbf{LSTM} & 10.4293 & 21.1792 & \multicolumn{1}{c}{11.6337} & 17.2426 & 15.1212 & 14.6796 & 35.4529 & \multicolumn{1}{c}{19.2637} & 27.4013 & 24.1994 \\
\textbf{GCRN}\cite{seo2018structured} & 9.7909 & 20.3683 &  \multicolumn{1}{c}{10.3356} & 15.8446 & 14.0849 & 13.1624 & 30.2009 &  \multicolumn{1}{c}{15.4056} & 21.6866 & 20.1139 \\
\textbf{LSTnet}\cite{lai2018modeling} & \textbf{9.4686} & 20.9138 & \multicolumn{1}{c}{10.3971} & 16.9684 & 14.4370 & \textbf{12.7202} & 33.5423 & \multicolumn{1}{c}{16.5796} & 27.1846 & 22.5067 \\
\textbf{DA-RNN}\cite{qin2017dual} & 10.7181 & 23.8095 & \multicolumn{1}{c}{11.5292} & 16.3595 & 15.6041 & 14.2535 & 33.8811 & \multicolumn{1}{c}{17.9019} & 24.0820 & 22.5296 \\ \hline
\textbf{MT-LSTM} & 10.2757 & 21.0571 & \multicolumn{1}{c}{11.2311} & 17.0639 & 14.9069 & 14.3336 & 33.4450 & \multicolumn{1}{c}{18.7191} & 27.6503 & 23.5370 \\
\textbf{MATURE} & 10.0121 & \textbf{19.9445} & \multicolumn{1}{c}{\textbf{9.6914}} & \textbf{15.3570} & \textbf{13.7512} & 13.9679 & \textbf{30.1061} & \multicolumn{1}{c}{\textbf{14.4222}} & \textbf{21.5486} & \textbf{20.0112} \\ \hline
\end{tabular}
\label{table1}
\end{table*}

\subsection{Performance Comparison} \label{sec:overall comparison}

For the tested approaches, we tune the model parameters on the validation dataset to locate the best parameters and list the forecasting results based on the testing dataset. A summary of the results of different models is reported in Table~\ref{table1}. Due to our aim is to enhance the forecasting accuracy of station-sparse sources with the help of useful knowledge learned from the station-intensive source, we list the MAE and RMSE of the three station-sparse sources (train, light rail, and ferry) and calculate the average values of the forecasting results of the bus in three experiments for testing three station-sparse sources for the basic multi-task model \textbf{MT-LSTM} and our model \textbf{MATURE}.

From Table~\ref{table1}, we make several observations. First, the classical machine learning models including HA and LR perform worse than other techniques in terms of both MAE and RMSE. Also, it seems that XGBoost has an advantage in some sources (e.g., train) for having a better performance than some deep learning methods. Then, in deep learning methods, MLP has a relatively poor performance since it is hard to extract the characteristics of temporal relationships. Other tested strategies based on recurrent neural networks (e.g., LSTM) or convolutional networks have the ability to handle this problem and thus improve forecasting accuracy. The three listed state-of-the-art strategies (GCRN, LSTnet, and DA-RNN) obtain better results than classic LSTM except for DA-RNN for the train and bus since it is proposed for uni-variant forecasting. It could predict accurately for those with fewer variants but has a poor ability for abundant variants prediction. The performance of GCRN is better than other state-of-the-art strategies and such results imply that capturing spatial and temporal features simultaneously does improve the forecasting results. This motivates us to consider the memory-based spatial-temporal correlations for demand forecasting in our future work.

Moreover, \textbf{MT-LSTM} yields better results than LSTM which means different sources indeed have connections. It is effective to concatenate several sources to find their possible relations and share the extracted knowledge for further prediction. However, LSTM is not as powerful as other state-of-the-art methods, so the MAE and RMSE values obtained by \textbf{MT-LSTM} are larger than them. Compared to \textbf{MT-LSTM}, our method gains $7.75\%$ relative improvement in MAE and $14.98\%$ relative improvement in RMSE on the average of the four modes which proves the effectiveness of the external memory module and the knowledge adaption strategy. Though, our method gains slightly larger MAE and RMSE on the bus than GCRN and LSTnet, jointly training the sources by \textbf{MATURE} obtains the highest accuracy regarding all the metrics in the three station-sparse sources. Also, compared to the best state-of-the-art method, our method obtains improvement in MAE and RMSE on the average of the four modes. Such results indicate that in the same city, based on our multi-task learning framework, extracting information from the station-intensive source to characterize the demand patterns of the target areas by the external memory-augmented recurrent network introduced in Section~\ref{sec:memory} and adapting the useful knowledge to station-sparse sources by the knowledge adaption module discussed in Section~\ref{sec:update} could enhance the forecasting performance of station-sparse sources. Also, the bus still achieves good performance which means our method will not destroy the temporal correlations of the station-intensive source.

\subsection{Ablation Study on Model Architecture}

To study the effects of different components of the proposed model, we further evaluate the models with various combinations of components. Table \ref{table2} lists the comparison results of our method and five different combinations. We only list the MAE and RMSE of the three station-sparse sources (train, ferry, and light rail) since our aim is to verify the demand patterns extracted from the station-intensive source has the ability to enhance the performance of station-sparse sources. The tested architectures are described as follows:

\begin{itemize}
    \item \textbf{C-LSTM:} The concatenation of the two sources is sent into the LSTM layer. And then two fully connected layers analyze the obtained matrices for demand prediction.
    \item \textbf{MT-LSTM:} As described in Section~\ref{sec:baseline}, two LSTM layers are utilized to extract temporal correlations for two sources separately. The extracted knowledge is concatenated and then sent into two fully connected layers for further prediction.
    \item \textbf{MARN:} The strategy we described in Section~\ref{sec:memory}. We train the model for predicting the demand of each public transportation mode independently by \textbf{MARN}.
    \item \textbf{MARN-S:} This architecture adopts \textbf{MARN} to extract temporal correlations for each source at first. Then, the hidden states are combined and followed by the fully connected layers for further prediction.
    \item \textbf{MARN-C:} This architecture also adopts \textbf{MARN} to extract knowledge from the sources independently. The procedure of knowledge adaption for the station-sparse source is to concatenate two external memories. The concatenation matrix is then sent to a fully connected layer to construct a new memory module $\mathbf{M}_{t}^{S}$ for station-sparse source. Also, the hidden states of two sources are combined together and sent into fully connected layers to predict the demand. 
\end{itemize}

\begin{table}[h]
\caption{Comparison with Different Variants}
\setlength{\tabcolsep}{0.3mm}{
\begin{tabular}{l|lll|lll}
\hline
\multirow{2}{*}{\textbf{Method}} & \multicolumn{3}{c|}{\textbf{MAE}} & \multicolumn{3}{c}{\textbf{RMSE}} \\ \cline{2-7} 
 & \textbf{Train} & \multicolumn{1}{c}{\textbf{\begin{tabular}[c]{@{}c@{}}Light \\ Rail\end{tabular}}} & \textbf{Ferry} & \textbf{Train} & \multicolumn{1}{c}{\textbf{\begin{tabular}[c]{@{}c@{}}Light \\ Rail\end{tabular}}} & \textbf{Ferry} \\ \hline
\textbf{C-LSTM} & 24.7948 & 11.9604 & 20.4253 & 38.2210 & 19.9569 & 33.3003 \\
\textbf{MT-LSTM} & 21.0571 & 11.2311 & 17.0639 & 33.4450 & 18.7191 & 27.6503 \\
\textbf{MARN} & 21.1013 & 11.5057 & 16.6581 & 31.6033 & 18.1840 & 24.7556 \\
\textbf{MARN-S} & 20.5172 & 11.4115 & 16.5313 & 30.3728 & 18.0119 & 24.1141\\
\textbf{MARN-C} & 20.3309 & 11.4738 & 16.2676 & 30.1797 & 18.0624 & 24.0286 \\
\textbf{Our Model} & \textbf{19.9445} & \textbf{9.6914} & \textbf{15.3570} & \textbf{30.1061} & \textbf{14.4222} & \textbf{21.5486} \\ \hline
\end{tabular}}
\label{table2}
\end{table}

\begin{figure*}[]
\centering
\subfigure[Train]{
\label{fig:train}
\includegraphics[width=.32\linewidth]{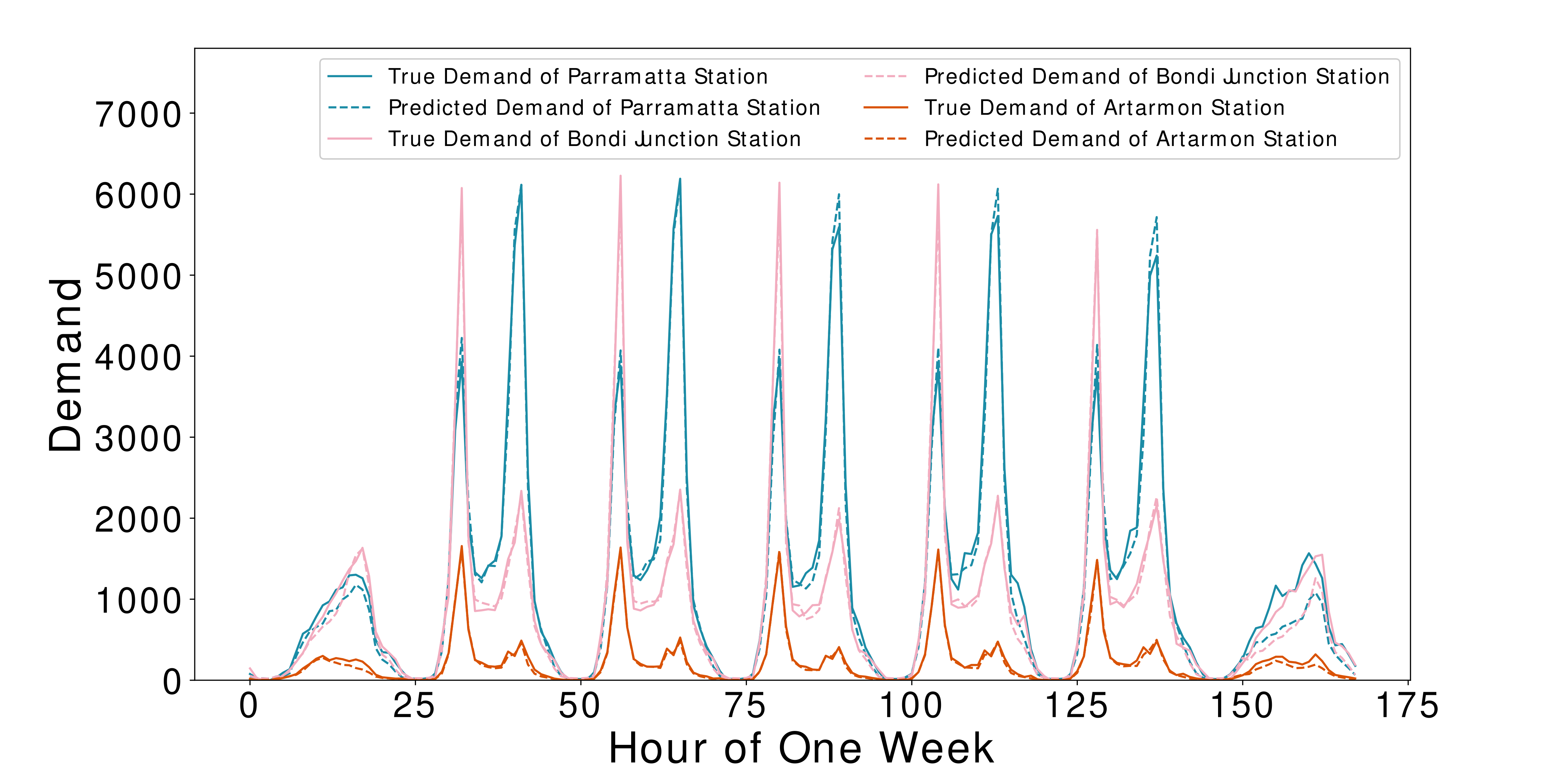}}
\subfigure[Light Rail]{
\label{fig:lr} 
\includegraphics[width=.32\linewidth]{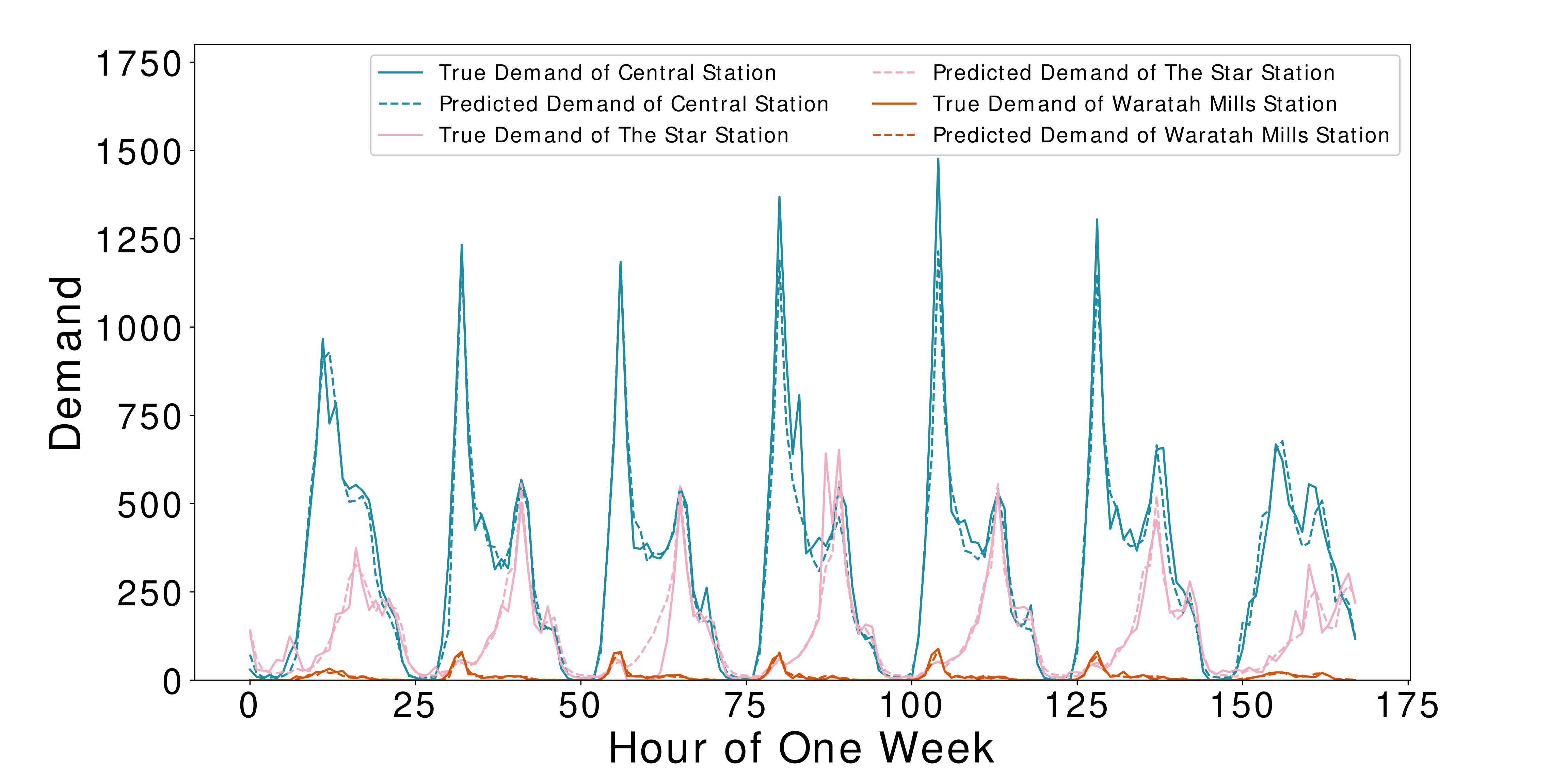}}
\subfigure[Ferry]{
\label{fig:ferry} 
\includegraphics[width=.32\linewidth]{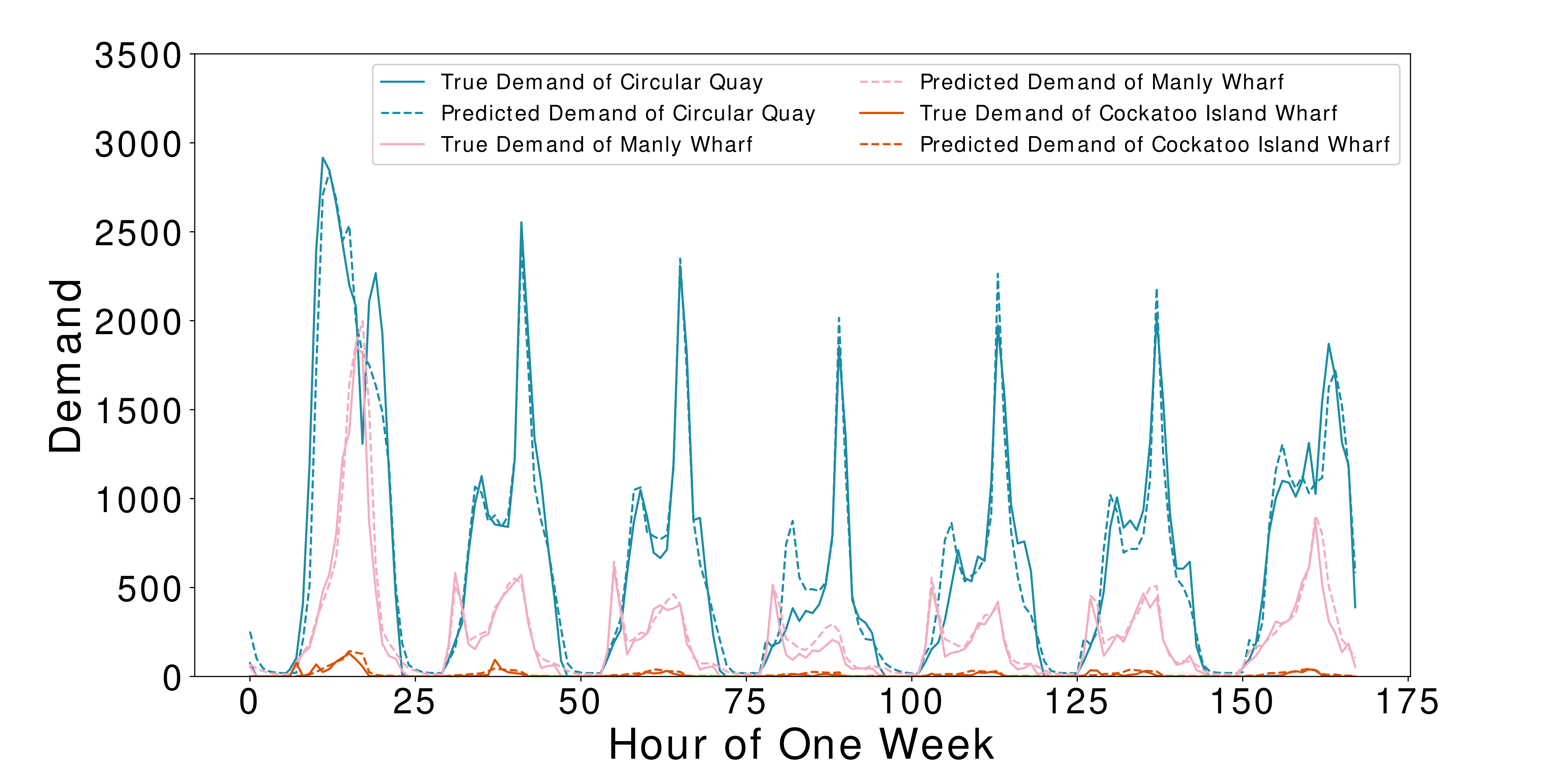}}
\caption{Forecasting Demand vs. True Demand}
\label{fig:example_result} 
\end{figure*}

\begin{figure*}[]
\centering
\subfigure[Train]{
\label{fig:train_para}
\includegraphics[width=.32\linewidth]{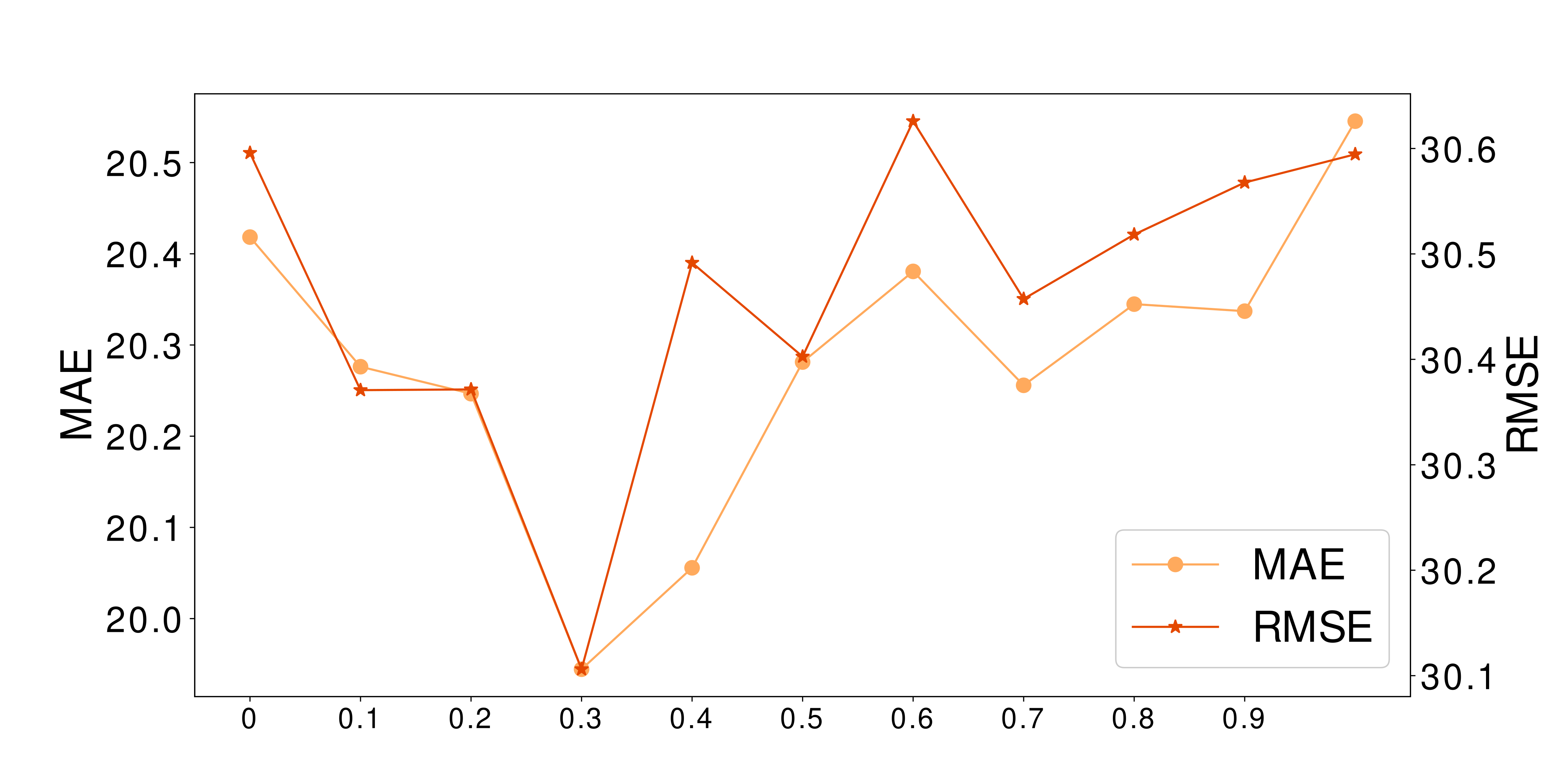}}
\subfigure[Light Rail]{
\label{fig:lr_para} 
\includegraphics[width=.32\linewidth]{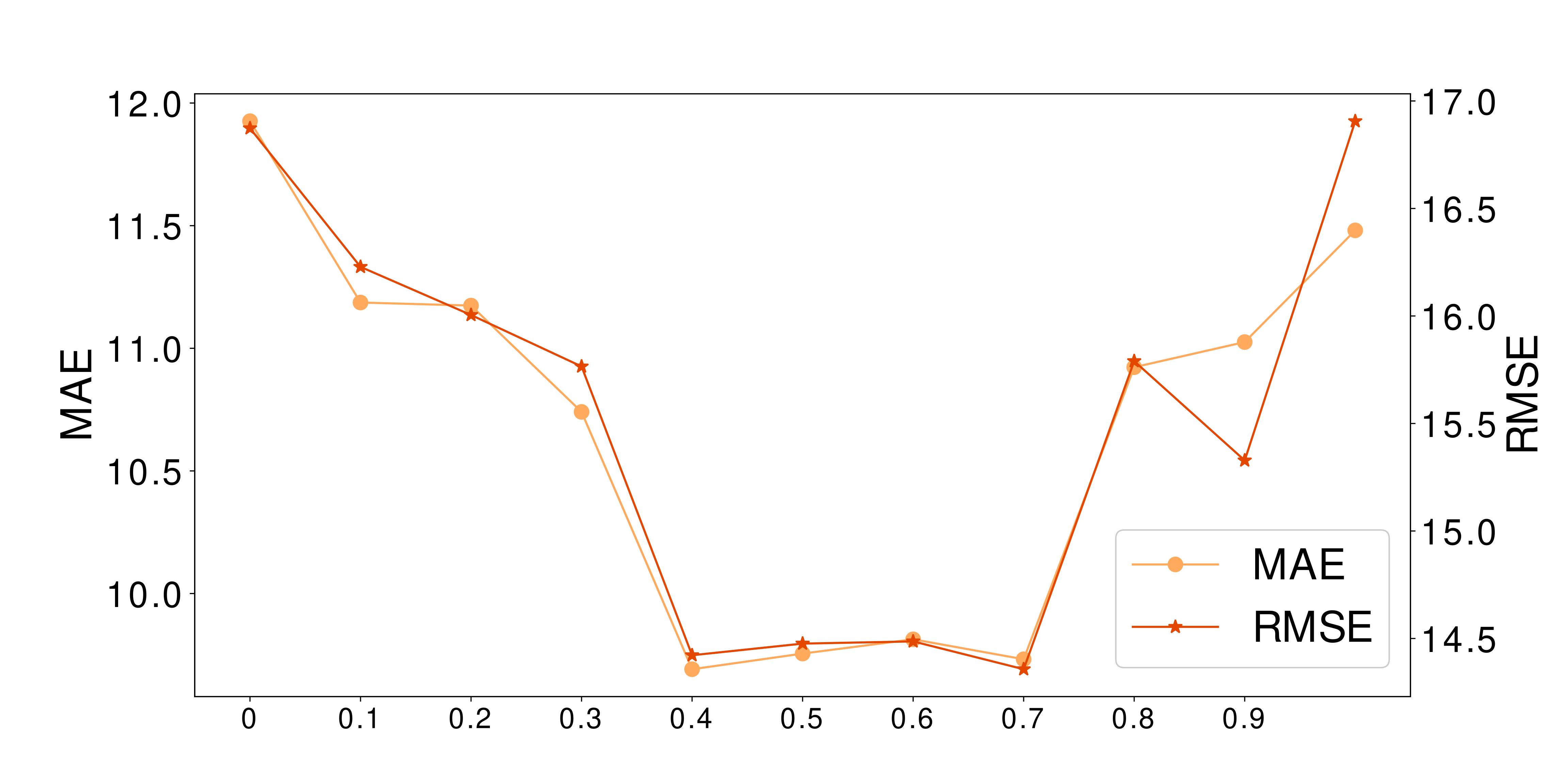}}
\subfigure[Ferry]{
\label{fig:ferry_para} 
\includegraphics[width=.32\linewidth]{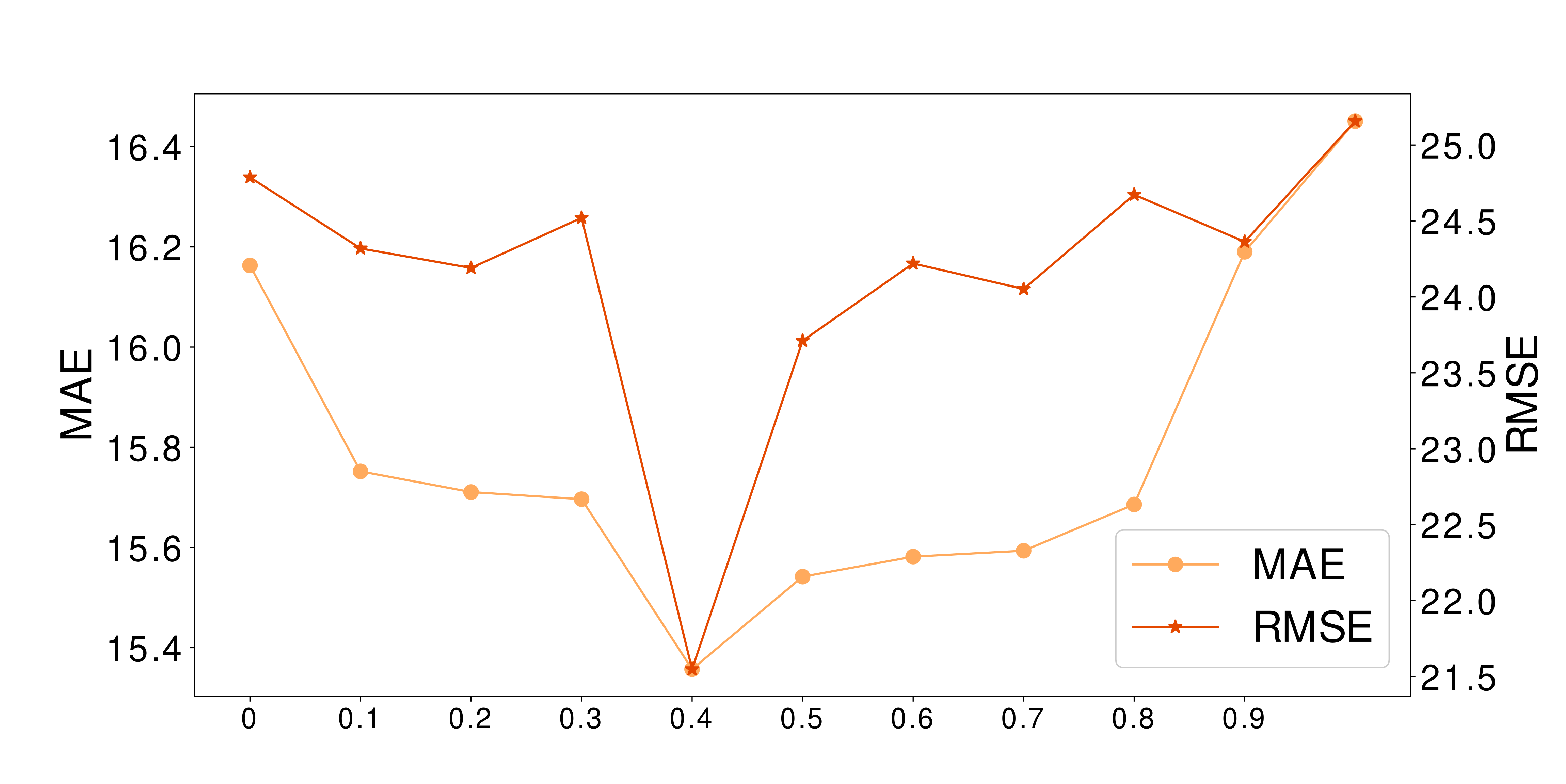}}
\caption{Parameter Sensitivity of $\gamma$}
\label{fig:para} 
\end{figure*}

From the forecasting results in Table~\ref{table2}, we can find that the first combination of LSTM, \textbf{C-LSTM} has worse accuracy than LSTM which means the concatenation of two sources may destroy the temporal correlations of the original data which will make a bad effect on demand forecasting. In contrast, the second structure \textbf{MT-LSTM} improves the prediction performance based on LSTM. These imply to us that it is better to analyze temporal relations at first for each mode and then find their underlying relations which could improve the results. Such experiments motivate us to take advantage of the station-intensive source to enhance the performance of station-sparse sources in our work. \textbf{MARN} achieves better performance than LSTM on the three sources, which illustrates \textbf{MARN} could enhance the capability to model long-and-short term information for prediction. The strategy \textbf{MARN-S} based on the external augmented module yields better performance on demand prediction than \textbf{MARN}, which indicates that analyzing the implicit correlations among the temporal information extracted from two sources can improve the forecasting accuracy. Though \textbf{MARN-S} gains larger MAE than \textbf{MT-LSTM} on the light rail, it has lower RMSE which means \textbf{MARN-S} works better on larger demand values. Then, \textbf{MARN-C} with external memory adaption obtains lower MAE and RMSE on the train and ferry which implies the significance of adapting information stored in memory $\mathbf{M}_{t}$. However, it achieves worse performance on the light rail and such fact implies that pure concatenation operation could bring noises to the station-sparse sources to influence the performance. Furthermore, our model achieves a better performance than the listed strategies which indicates that the knowledge from the station-intensive source can play a strengthening effect on forecasting the demand of station-sparse sources. And it demonstrates our knowledge adaption module is more powerful to adapt knowledge from $\mathbf{M}_{t}^{R}$ to $\mathbf{M}_{t}^{S}$ than pure concatenation which is hard to avoid meaningless specific features. The proposed method is able to effectively decrease the impact of noise and increase the impact of useful knowledge from the station-intensive source on station-sparse sources. 

\subsection{Comparison of Different Stations}

We now move to illustrate the effectiveness of the proposed method on the selected stations with different ranges of demand. 

Specifically, we choose three stations for the three station-sparse sources to predict the demand. The selected stations have totally different average values and standard deviation distribution of passenger demand. We plot the true demand and prediction results obtained by our model in Figure~\ref{fig:example_result} where the solid lines denote true data while the dashed lines denote forecasting demand. In general, the proposed model \textbf{MATURE} has the ability to predict the demand of each hour for different ranges of demand precisely. The predicted curves and the real curves are highly coincident. As for the train, our model can accurately predict peaks except for three occasions where demand values are smaller than those of the previous two days and are expected to be irregular. The predicted demand on the last day is less accurate than other days since there exists a drop that is hard to be detected. For the light rail, there also exist several peaks which are not fully captured because of the drop. The ferry demand is often affected by weather disturbances. Thus, the accuracy of the ferry is slightly lower than the other two public transit modes. For instance, the demand of the fourth day is much lower than the other six days which is an irregular case that is hard to be predicted (without additional inputs or information). Such phenomenons indeed motivate us to focus on the demand peak and drop capturing in our future work. In summary, the proposed method can predict the demand of the ferry accurately except for cases that are largely affected by other external factors.

\subsection{Parameter Sensitivity} \label{sec:parameter}
To study the influence of hyperparameters of \textbf{MATURE}, we then train the proposed model on our dataset with different values of $\gamma$ in Formula~(\ref{formula:12}).

Specifically, $\gamma$ is changed from $0$ to $1$ when fixing the value of the learning rate and $\epsilon$ in the loss function. Figure \ref{fig:para} summarizes the results where the left x-axis represents MAE while the right x-axis represents RMSE. When we set $\gamma$ to $0$, the adaptive function only contains the matrix $\mathbf{M}_{t}^{new}$ that we calculate through the knowledge adaption module. When we set $\gamma$ to $1$, the adaptive procedure is invalid and the method is similar to the model \textbf{MARN-S}. The results also match with the MAE and RMSE of \textbf{MARN-S}. As the value of $\gamma$ changes, the values of MAE do not fluctuate significantly, indicating that our model is not sensitive to this hyperparameter. Overall, $\gamma$ equaling to $0.3$, $0.4$, and $0.3$ obtains the highest accuracy for the train, light rail, and ferry. The above results indicate that the knowledge adaption mechanism is impressive to adapt information from the station-intensive source to the station-sparse source for predicting enhancement.

\section{Conclusion} \label{sec:conclusion}
In this paper, we propose a novel external memory-based multi-task model, namely Memory-Augmented Multi-task Recurrent Network (\textbf{MATURE}) for demand forecasting regarding station-sparse public transit modes with the help of station-intensive mode(s). Specifically, the method learns an external memory based on the recurrent network for each data source including station-intensive mode and station-sparse modes which can strengthen the ability to capture the temporal information and store the useful knowledge for further prediction. The knowledge extracted from the station-intensive source often captures the demand patterns/features of the selected areas well. Thus, we then introduce a knowledge adaption module to adapt the knowledge from the station-intensive source to station-sparse sources which could enhance the predicting performance of station-sparse public transit modes. When evaluated on one real-world dataset including four public transit modes (bus, train, light rail, and ferry), our approach achieves better performance than the state-of-the-art baselines. This research provides a new tool and insights to the study of public transport demand prediction by extracting knowledge from various modes and sharing/adapting useful features to the modes that need them. For further work, we will explore the spatial correlations and spatio-temporal relations for demand prediction based on our model.

\bibliographystyle{ACM-Reference-Format}
\bibliography{acmart}


\begin{thebibliography}{31}


\ifx \showCODEN    \undefined \def \showCODEN     #1{\unskip}     \fi
\ifx \showDOI      \undefined \def \showDOI       #1{#1}\fi
\ifx \showISBNx    \undefined \def \showISBNx     #1{\unskip}     \fi
\ifx \showISBNxiii \undefined \def \showISBNxiii  #1{\unskip}     \fi
\ifx \showISSN     \undefined \def \showISSN      #1{\unskip}     \fi
\ifx \showLCCN     \undefined \def \showLCCN      #1{\unskip}     \fi
\ifx \shownote     \undefined \def \shownote      #1{#1}          \fi
\ifx \showarticletitle \undefined \def \showarticletitle #1{#1}   \fi
\ifx \showURL      \undefined \def \showURL       {\relax}        \fi
\providecommand\bibfield[2]{#2}
\providecommand\bibinfo[2]{#2}
\providecommand\natexlab[1]{#1}
\providecommand\showeprint[2][]{arXiv:#2}

\bibitem[\protect\citeauthoryear{Bai, Yao, Kanhere, Wang, Liu, and Yang}{Bai
  et~al\mbox{.}}{2019b}]%
        {bai2019spatio}
\bibfield{author}{\bibinfo{person}{Lei Bai}, \bibinfo{person}{Lina Yao},
  \bibinfo{person}{Salil~S Kanhere}, \bibinfo{person}{Xianzhi Wang},
  \bibinfo{person}{Wei Liu}, {and} \bibinfo{person}{Zheng Yang}.}
  \bibinfo{year}{2019}\natexlab{b}.
\newblock \showarticletitle{Spatio-Temporal Graph Convolutional and Recurrent
  Networks for Citywide Passenger Demand Prediction}. In
  \bibinfo{booktitle}{\emph{Proceedings of the 28th ACM International
  Conference on Information and Knowledge Management}}.
  \bibinfo{pages}{2293--2296}.
\newblock


\bibitem[\protect\citeauthoryear{Bai, Yao, Kanhere, Wang, and Sheng}{Bai
  et~al\mbox{.}}{2019a}]%
        {bai2019stg2seq}
\bibfield{author}{\bibinfo{person}{Lei Bai}, \bibinfo{person}{Lina Yao},
  \bibinfo{person}{Salil~S Kanhere}, \bibinfo{person}{Xianzhi Wang}, {and}
  \bibinfo{person}{Quan~Z Sheng}.} \bibinfo{year}{2019}\natexlab{a}.
\newblock \showarticletitle{STG2seq: spatial-temporal graph to sequence model
  for multi-step passenger demand forecasting}. In
  \bibinfo{booktitle}{\emph{Proceedings of the 28th International Joint
  Conference on Artificial Intelligence}}. AAAI Press,
  \bibinfo{pages}{1981--1987}.
\newblock


\bibitem[\protect\citeauthoryear{Bai, Yao, Kanhere, Yang, Chu, and Wang}{Bai
  et~al\mbox{.}}{2019c}]%
        {bai2019passenger}
\bibfield{author}{\bibinfo{person}{Lei Bai}, \bibinfo{person}{Lina Yao},
  \bibinfo{person}{Salil~S Kanhere}, \bibinfo{person}{Zheng Yang},
  \bibinfo{person}{Jing Chu}, {and} \bibinfo{person}{Xianzhi Wang}.}
  \bibinfo{year}{2019}\natexlab{c}.
\newblock \showarticletitle{Passenger demand forecasting with multi-task
  convolutional recurrent neural networks}. In
  \bibinfo{booktitle}{\emph{Pacific-Asia Conference on Knowledge Discovery and
  Data Mining}}. Springer, \bibinfo{pages}{29--42}.
\newblock


\bibitem[\protect\citeauthoryear{Chen and Guestrin}{Chen and Guestrin}{2016}]%
        {chen2016xgboost}
\bibfield{author}{\bibinfo{person}{Tianqi Chen} {and} \bibinfo{person}{Carlos
  Guestrin}.} \bibinfo{year}{2016}\natexlab{}.
\newblock \showarticletitle{Xgboost: A scalable tree boosting system}. In
  \bibinfo{booktitle}{\emph{Proceedings of the 22nd ACM SIGKDD International
  Conference on Knowledge Discovery and Data Mining}}. ACM,
  \bibinfo{pages}{785--794}.
\newblock


\bibitem[\protect\citeauthoryear{Cirstea, Micu, Muresan, Guo, and Yang}{Cirstea
  et~al\mbox{.}}{2018}]%
        {cirstea2018correlated}
\bibfield{author}{\bibinfo{person}{Razvan-Gabriel Cirstea},
  \bibinfo{person}{Darius-Valer Micu}, \bibinfo{person}{Gabriel-Marcel
  Muresan}, \bibinfo{person}{Chenjuan Guo}, {and} \bibinfo{person}{Bin Yang}.}
  \bibinfo{year}{2018}\natexlab{}.
\newblock \showarticletitle{Correlated time series forecasting using multi-task
  deep neural networks}. In \bibinfo{booktitle}{\emph{Proceedings of the 27th
  acm international conference on information and knowledge management}}.
  \bibinfo{pages}{1527--1530}.
\newblock


\bibitem[\protect\citeauthoryear{Hochreiter and Schmidhuber}{Hochreiter and
  Schmidhuber}{1997}]%
        {hochreiter1997long}
\bibfield{author}{\bibinfo{person}{Sepp Hochreiter} {and}
  \bibinfo{person}{J{\"u}rgen Schmidhuber}.} \bibinfo{year}{1997}\natexlab{}.
\newblock \showarticletitle{Long short-term memory}.
\newblock \bibinfo{journal}{\emph{Neural computation}} \bibinfo{volume}{9},
  \bibinfo{number}{8} (\bibinfo{year}{1997}), \bibinfo{pages}{1735--1780}.
\newblock


\bibitem[\protect\citeauthoryear{Huang, Wang, Wu, and Tang}{Huang
  et~al\mbox{.}}{2019}]%
        {huang2019dsanet}
\bibfield{author}{\bibinfo{person}{Siteng Huang}, \bibinfo{person}{Donglin
  Wang}, \bibinfo{person}{Xuehan Wu}, {and} \bibinfo{person}{Ao Tang}.}
  \bibinfo{year}{2019}\natexlab{}.
\newblock \showarticletitle{DSANet: Dual Self-Attention Network for
  Multivariate Time Series Forecasting}. In
  \bibinfo{booktitle}{\emph{Proceedings of the 28th ACM International
  Conference on Information and Knowledge Management}}.
  \bibinfo{pages}{2129--2132}.
\newblock


\bibitem[\protect\citeauthoryear{Kipf and Welling}{Kipf and Welling}{2016}]%
        {kipf2016semi}
\bibfield{author}{\bibinfo{person}{Thomas~N Kipf} {and} \bibinfo{person}{Max
  Welling}.} \bibinfo{year}{2016}\natexlab{}.
\newblock \showarticletitle{Semi-supervised classification with graph
  convolutional networks}.
\newblock \bibinfo{journal}{\emph{arXiv preprint arXiv:1609.02907}}
  (\bibinfo{year}{2016}).
\newblock


\bibitem[\protect\citeauthoryear{Lai, Chang, Yang, and Liu}{Lai
  et~al\mbox{.}}{2018}]%
        {lai2018modeling}
\bibfield{author}{\bibinfo{person}{Guokun Lai}, \bibinfo{person}{Wei-Cheng
  Chang}, \bibinfo{person}{Yiming Yang}, {and} \bibinfo{person}{Hanxiao Liu}.}
  \bibinfo{year}{2018}\natexlab{}.
\newblock \showarticletitle{Modeling long-and short-term temporal patterns with
  deep neural networks}. In \bibinfo{booktitle}{\emph{The 41st International
  ACM SIGIR Conference on Research \& Development in Information Retrieval}}.
  ACM, \bibinfo{pages}{95--104}.
\newblock


\bibitem[\protect\citeauthoryear{LeCun, Bottou, Bengio, and Haffner}{LeCun
  et~al\mbox{.}}{1998}]%
        {lecun1998gradient}
\bibfield{author}{\bibinfo{person}{Yann LeCun}, \bibinfo{person}{L{\'e}on
  Bottou}, \bibinfo{person}{Yoshua Bengio}, {and} \bibinfo{person}{Patrick
  Haffner}.} \bibinfo{year}{1998}\natexlab{}.
\newblock \showarticletitle{Gradient-based learning applied to document
  recognition}.
\newblock \bibinfo{journal}{\emph{Proc. IEEE}} \bibinfo{volume}{86},
  \bibinfo{number}{11} (\bibinfo{year}{1998}), \bibinfo{pages}{2278--2324}.
\newblock


\bibitem[\protect\citeauthoryear{Li, Yu, Shahabi, and Liu}{Li
  et~al\mbox{.}}{2018}]%
        {li2017diffusion}
\bibfield{author}{\bibinfo{person}{Yaguang Li}, \bibinfo{person}{Rose Yu},
  \bibinfo{person}{Cyrus Shahabi}, {and} \bibinfo{person}{Yan Liu}.}
  \bibinfo{year}{2018}\natexlab{}.
\newblock \showarticletitle{Diffusion Convolutional Recurrent Neural Network:
  Data-Driven Traffic Forecasting}. In \bibinfo{booktitle}{\emph{International
  Conference on Learning Representations}}.
\newblock


\bibitem[\protect\citeauthoryear{Lippi, Bertini, and Frasconi}{Lippi
  et~al\mbox{.}}{2013}]%
        {lippi2013short}
\bibfield{author}{\bibinfo{person}{Marco Lippi}, \bibinfo{person}{Matteo
  Bertini}, {and} \bibinfo{person}{Paolo Frasconi}.}
  \bibinfo{year}{2013}\natexlab{}.
\newblock \showarticletitle{Short-term traffic flow forecasting: An
  experimental comparison of time-series analysis and supervised learning}.
\newblock \bibinfo{journal}{\emph{IEEE Transactions on Intelligent
  Transportation Systems}} \bibinfo{volume}{14}, \bibinfo{number}{2}
  (\bibinfo{year}{2013}), \bibinfo{pages}{871--882}.
\newblock


\bibitem[\protect\citeauthoryear{Liu, Qiu, and Huang}{Liu
  et~al\mbox{.}}{2016a}]%
        {liu2016recurrent}
\bibfield{author}{\bibinfo{person}{Pengfei Liu}, \bibinfo{person}{Xipeng Qiu},
  {and} \bibinfo{person}{Xuanjing Huang}.} \bibinfo{year}{2016}\natexlab{a}.
\newblock \showarticletitle{Recurrent neural network for text classification
  with multi-task learning}. In \bibinfo{booktitle}{\emph{Proceedings of the
  Twenty-Fifth International Joint Conference on Artificial Intelligence}}.
  \bibinfo{pages}{2873--2879}.
\newblock


\bibitem[\protect\citeauthoryear{Liu, Qiu, and Huang}{Liu
  et~al\mbox{.}}{2016b}]%
        {liu2016deep}
\bibfield{author}{\bibinfo{person}{Pengfei Liu}, \bibinfo{person}{Xipeng Qiu},
  {and} \bibinfo{person}{Xuan-Jing Huang}.} \bibinfo{year}{2016}\natexlab{b}.
\newblock \showarticletitle{Deep Multi-Task Learning with Shared Memory for
  Text Classification}. In \bibinfo{booktitle}{\emph{Proceedings of the 2016
  Conference on Empirical Methods in Natural Language Processing}}.
  \bibinfo{pages}{118--127}.
\newblock


\bibitem[\protect\citeauthoryear{Moreira-Matias, Gama, Ferreira,
  Mendes-Moreira, and Damas}{Moreira-Matias et~al\mbox{.}}{2013}]%
        {moreira2013predicting}
\bibfield{author}{\bibinfo{person}{Luis Moreira-Matias}, \bibinfo{person}{Joao
  Gama}, \bibinfo{person}{Michel Ferreira}, \bibinfo{person}{Joao
  Mendes-Moreira}, {and} \bibinfo{person}{Luis Damas}.}
  \bibinfo{year}{2013}\natexlab{}.
\newblock \showarticletitle{Predicting taxi--passenger demand using streaming
  data}.
\newblock \bibinfo{journal}{\emph{IEEE Transactions on Intelligent
  Transportation Systems}} \bibinfo{volume}{14}, \bibinfo{number}{3}
  (\bibinfo{year}{2013}), \bibinfo{pages}{1393--1402}.
\newblock


\bibitem[\protect\citeauthoryear{Qi, Li, Deng, Cai, Qi, and Deng}{Qi
  et~al\mbox{.}}{2019}]%
        {qi2019deep}
\bibfield{author}{\bibinfo{person}{Yan Qi}, \bibinfo{person}{Chenliang Li},
  \bibinfo{person}{Han Deng}, \bibinfo{person}{Min Cai},
  \bibinfo{person}{Yunwei Qi}, {and} \bibinfo{person}{Yuming Deng}.}
  \bibinfo{year}{2019}\natexlab{}.
\newblock \showarticletitle{A Deep Neural Framework for Sales Forecasting in
  E-Commerce}. In \bibinfo{booktitle}{\emph{Proceedings of the 28th ACM
  International Conference on Information and Knowledge Management}}.
  \bibinfo{pages}{299--308}.
\newblock


\bibitem[\protect\citeauthoryear{Qin, Song, Cheng, Cheng, Jiang, and
  Cottrell}{Qin et~al\mbox{.}}{2017}]%
        {qin2017dual}
\bibfield{author}{\bibinfo{person}{Yao Qin}, \bibinfo{person}{Dongjin Song},
  \bibinfo{person}{Haifeng Cheng}, \bibinfo{person}{Wei Cheng},
  \bibinfo{person}{Guofei Jiang}, {and} \bibinfo{person}{Garrison~W Cottrell}.}
  \bibinfo{year}{2017}\natexlab{}.
\newblock \showarticletitle{A dual-stage attention-based recurrent neural
  network for time series prediction}. In \bibinfo{booktitle}{\emph{Proceedings
  of the 26th International Joint Conference on Artificial Intelligence}}.
  \bibinfo{pages}{2627--2633}.
\newblock


\bibitem[\protect\citeauthoryear{Rae, Hunt, Danihelka, Harley, Senior, Wayne,
  Graves, and Lillicrap}{Rae et~al\mbox{.}}{2016}]%
        {rae2016scaling}
\bibfield{author}{\bibinfo{person}{Jack Rae}, \bibinfo{person}{Jonathan~J
  Hunt}, \bibinfo{person}{Ivo Danihelka}, \bibinfo{person}{Timothy Harley},
  \bibinfo{person}{Andrew~W Senior}, \bibinfo{person}{Gregory Wayne},
  \bibinfo{person}{Alex Graves}, {and} \bibinfo{person}{Timothy Lillicrap}.}
  \bibinfo{year}{2016}\natexlab{}.
\newblock \showarticletitle{Scaling memory-augmented neural networks with
  sparse reads and writes}. In \bibinfo{booktitle}{\emph{Advances in Neural
  Information Processing Systems}}. \bibinfo{pages}{3621--3629}.
\newblock


\bibitem[\protect\citeauthoryear{Seo, Defferrard, Vandergheynst, and
  Bresson}{Seo et~al\mbox{.}}{2018}]%
        {seo2018structured}
\bibfield{author}{\bibinfo{person}{Youngjoo Seo}, \bibinfo{person}{Micha{\"e}l
  Defferrard}, \bibinfo{person}{Pierre Vandergheynst}, {and}
  \bibinfo{person}{Xavier Bresson}.} \bibinfo{year}{2018}\natexlab{}.
\newblock \showarticletitle{Structured sequence modeling with graph
  convolutional recurrent networks}. In \bibinfo{booktitle}{\emph{International
  Conference on Neural Information Processing}}. Springer,
  \bibinfo{pages}{362--373}.
\newblock


\bibitem[\protect\citeauthoryear{Wang, Geng, Ma, Liu, and Yang}{Wang
  et~al\mbox{.}}{2019a}]%
        {wang2019cross}
\bibfield{author}{\bibinfo{person}{Leye Wang}, \bibinfo{person}{Xu Geng},
  \bibinfo{person}{Xiaojuan Ma}, \bibinfo{person}{Feng Liu}, {and}
  \bibinfo{person}{Qiang Yang}.} \bibinfo{year}{2019}\natexlab{a}.
\newblock \showarticletitle{Cross-city Transfer Learning for Deep
  Spatio-temporal Prediction}. In \bibinfo{booktitle}{\emph{IJCAI International
  Joint Conference on Artificial Intelligence}}. \bibinfo{pages}{1893}.
\newblock


\bibitem[\protect\citeauthoryear{Wang, Yin, Chen, Wo, Xu, and Zheng}{Wang
  et~al\mbox{.}}{2019b}]%
        {wang2019origin}
\bibfield{author}{\bibinfo{person}{Yuandong Wang}, \bibinfo{person}{Hongzhi
  Yin}, \bibinfo{person}{Hongxu Chen}, \bibinfo{person}{Tianyu Wo},
  \bibinfo{person}{Jie Xu}, {and} \bibinfo{person}{Kai Zheng}.}
  \bibinfo{year}{2019}\natexlab{b}.
\newblock \showarticletitle{Origin-destination matrix prediction via graph
  convolution: a new perspective of passenger demand modeling}. In
  \bibinfo{booktitle}{\emph{Proceedings of the 25th ACM SIGKDD International
  Conference on Knowledge Discovery \& Data Mining}}. ACM,
  \bibinfo{pages}{1227--1235}.
\newblock


\bibitem[\protect\citeauthoryear{Wang, Zhang, Zhu, Long, Wang, and Yu}{Wang
  et~al\mbox{.}}{2019c}]%
        {wang2019memory}
\bibfield{author}{\bibinfo{person}{Yunbo Wang}, \bibinfo{person}{Jianjin
  Zhang}, \bibinfo{person}{Hongyu Zhu}, \bibinfo{person}{Mingsheng Long},
  \bibinfo{person}{Jianmin Wang}, {and} \bibinfo{person}{Philip~S Yu}.}
  \bibinfo{year}{2019}\natexlab{c}.
\newblock \showarticletitle{Memory In Memory: A Predictive Neural Network for
  Learning Higher-Order Non-Stationarity from Spatiotemporal Dynamics}. In
  \bibinfo{booktitle}{\emph{Proceedings of the IEEE Conference on Computer
  Vision and Pattern Recognition}}. \bibinfo{pages}{9154--9162}.
\newblock


\bibitem[\protect\citeauthoryear{Wang, Fu, and Ye}{Wang et~al\mbox{.}}{2018}]%
        {wang2018learning}
\bibfield{author}{\bibinfo{person}{Zheng Wang}, \bibinfo{person}{Kun Fu}, {and}
  \bibinfo{person}{Jieping Ye}.} \bibinfo{year}{2018}\natexlab{}.
\newblock \showarticletitle{Learning to estimate the travel time}. In
  \bibinfo{booktitle}{\emph{Proceedings of the 24th ACM SIGKDD International
  Conference on Knowledge Discovery \& Data Mining}}.
  \bibinfo{pages}{858--866}.
\newblock


\bibitem[\protect\citeauthoryear{Wei, Zheng, and Yang}{Wei
  et~al\mbox{.}}{2016}]%
        {wei2016transfer}
\bibfield{author}{\bibinfo{person}{Ying Wei}, \bibinfo{person}{Yu Zheng}, {and}
  \bibinfo{person}{Qiang Yang}.} \bibinfo{year}{2016}\natexlab{}.
\newblock \showarticletitle{Transfer knowledge between cities}. In
  \bibinfo{booktitle}{\emph{Proceedings of the 22nd ACM SIGKDD International
  Conference on Knowledge Discovery and Data Mining}}.
  \bibinfo{pages}{1905--1914}.
\newblock


\bibitem[\protect\citeauthoryear{Xu, Rahmatizadeh, B{\"o}l{\"o}ni, and
  Turgut}{Xu et~al\mbox{.}}{2017}]%
        {xu2017real}
\bibfield{author}{\bibinfo{person}{Jun Xu}, \bibinfo{person}{Rouhollah
  Rahmatizadeh}, \bibinfo{person}{Ladislau B{\"o}l{\"o}ni}, {and}
  \bibinfo{person}{Damla Turgut}.} \bibinfo{year}{2017}\natexlab{}.
\newblock \showarticletitle{Real-time prediction of taxi demand using recurrent
  neural networks}.
\newblock \bibinfo{journal}{\emph{IEEE Transactions on Intelligent
  Transportation Systems}} \bibinfo{volume}{19}, \bibinfo{number}{8}
  (\bibinfo{year}{2017}), \bibinfo{pages}{2572--2581}.
\newblock


\bibitem[\protect\citeauthoryear{Xue, Sun, and Chen}{Xue et~al\mbox{.}}{2015}]%
        {xue2015short}
\bibfield{author}{\bibinfo{person}{Rui Xue}, \bibinfo{person}{Daniel~Jian Sun},
  {and} \bibinfo{person}{Shukai Chen}.} \bibinfo{year}{2015}\natexlab{}.
\newblock \showarticletitle{Short-term bus passenger demand prediction based on
  time series model and interactive multiple model approach}.
\newblock \bibinfo{journal}{\emph{Discrete Dynamics in Nature and Society}}
  \bibinfo{volume}{2015} (\bibinfo{year}{2015}).
\newblock


\bibitem[\protect\citeauthoryear{Yao, Liu, Wei, Tang, and Li}{Yao
  et~al\mbox{.}}{2019}]%
        {yao2019learning}
\bibfield{author}{\bibinfo{person}{Huaxiu Yao}, \bibinfo{person}{Yiding Liu},
  \bibinfo{person}{Ying Wei}, \bibinfo{person}{Xianfeng Tang}, {and}
  \bibinfo{person}{Zhenhui Li}.} \bibinfo{year}{2019}\natexlab{}.
\newblock \showarticletitle{Learning from multiple cities: A meta-learning
  approach for spatial-temporal prediction}. In \bibinfo{booktitle}{\emph{The
  World Wide Web Conference}}. \bibinfo{pages}{2181--2191}.
\newblock


\bibitem[\protect\citeauthoryear{Yao, Wu, Ke, Tang, Jia, Lu, Gong, Ye, and
  Li}{Yao et~al\mbox{.}}{2018}]%
        {yao2018deep}
\bibfield{author}{\bibinfo{person}{Huaxiu Yao}, \bibinfo{person}{Fei Wu},
  \bibinfo{person}{Jintao Ke}, \bibinfo{person}{Xianfeng Tang},
  \bibinfo{person}{Yitian Jia}, \bibinfo{person}{Siyu Lu},
  \bibinfo{person}{Pinghua Gong}, \bibinfo{person}{Jieping Ye}, {and}
  \bibinfo{person}{Zhenhui Li}.} \bibinfo{year}{2018}\natexlab{}.
\newblock \showarticletitle{Deep multi-view spatial-temporal network for taxi
  demand prediction}. In \bibinfo{booktitle}{\emph{Thirty-Second AAAI
  Conference on Artificial Intelligence}}.
\newblock


\bibitem[\protect\citeauthoryear{Ye, Sun, Du, Fu, Tong, and Xiong}{Ye
  et~al\mbox{.}}{2019}]%
        {ye2019co}
\bibfield{author}{\bibinfo{person}{Junchen Ye}, \bibinfo{person}{Leilei Sun},
  \bibinfo{person}{Bowen Du}, \bibinfo{person}{Yanjie Fu},
  \bibinfo{person}{Xinran Tong}, {and} \bibinfo{person}{Hui Xiong}.}
  \bibinfo{year}{2019}\natexlab{}.
\newblock \showarticletitle{Co-Prediction of Multiple Transportation Demands
  Based on Deep Spatio-Temporal Neural Network}. In
  \bibinfo{booktitle}{\emph{Proceedings of the 25th ACM SIGKDD International
  Conference on Knowledge Discovery \& Data Mining}}.
  \bibinfo{pages}{305--313}.
\newblock


\bibitem[\protect\citeauthoryear{Yi, Zhang, Wang, Li, and Zheng}{Yi
  et~al\mbox{.}}{2018}]%
        {yi2018deep}
\bibfield{author}{\bibinfo{person}{Xiuwen Yi}, \bibinfo{person}{Junbo Zhang},
  \bibinfo{person}{Zhaoyuan Wang}, \bibinfo{person}{Tianrui Li}, {and}
  \bibinfo{person}{Yu Zheng}.} \bibinfo{year}{2018}\natexlab{}.
\newblock \showarticletitle{Deep distributed fusion network for air quality
  prediction}. In \bibinfo{booktitle}{\emph{Proceedings of the 24th ACM SIGKDD
  International Conference on Knowledge Discovery \& Data Mining}}.
  \bibinfo{pages}{965--973}.
\newblock


\bibitem[\protect\citeauthoryear{Zhou, Shen, Zhu, and Huang}{Zhou
  et~al\mbox{.}}{2018}]%
        {zhou2018predicting}
\bibfield{author}{\bibinfo{person}{Xian Zhou}, \bibinfo{person}{Yanyan Shen},
  \bibinfo{person}{Yanmin Zhu}, {and} \bibinfo{person}{Linpeng Huang}.}
  \bibinfo{year}{2018}\natexlab{}.
\newblock \showarticletitle{Predicting multi-step citywide passenger demands
  using attention-based neural networks}. In
  \bibinfo{booktitle}{\emph{Proceedings of the Eleventh ACM International
  Conference on Web Search and Data Mining}}. \bibinfo{pages}{736--744}.
\newblock


\end{thebibliography}


\end{document}